%% file: main.tex
\documentclass[sigconf,nonacm]{acmart}

\AtBeginDocument{%
  }

\author{Viktor Muryn$^\dagger$ \quad Marta Sumyk$^\dagger$ \quad Mariya Hirna$^*$ \quad Sofiya Garkot$^*$ \quad Maksym Shamrai$^*$}

\affiliation{$^\dagger$Ukrainian Catholic University, Lviv \country{Ukraine}}
\affiliation{$^*$MacPaw, Kyiv \country{Ukraine} \\
\texttt{\{muryn.pn, sumyk.pn\}@ucu.edu.ua, \{maryhirna, sofiyagarkot, mshamrai\}@macpaw.com}
}

\usepackage{listings}
\usepackage{subcaption}



\title{Screen2AX: Vision-Based Approach for Automatic macOS Accessibility Generation}

\begin{document}

\begin{abstract}

Desktop accessibility metadata enables AI agents to interpret screens and supports users who depend on tools like screen readers. Yet, many applications remain largely inaccessible due to incomplete or missing metadata provided by developers—our investigation shows that only 33\% of applications on macOS offer full accessibility support. While recent work on structured screen representation has primarily addressed specific challenges, such as UI element detection or captioning, none has attempted to capture the full complexity of desktop interfaces by replicating their entire hierarchical structure. 

To bridge this gap, we introduce Screen2AX, the first framework to automatically create real-time, tree-structured accessibility metadata from a single screenshot. Our method uses vision-language and object detection models to detect, describe, and organize UI elements hierarchically, mirroring macOS’s system-level accessibility structure. 

To tackle the limited availability of data for macOS desktop applications, we compiled and publicly released three datasets encompassing 112 macOS applications, each annotated for UI element detection, grouping, and hierarchical accessibility metadata alongside corresponding screenshots. Screen2AX accurately infers hierarchy trees, achieving a 77\% F1 score in reconstructing a complete accessibility tree. Crucially, these hierarchy trees improve the ability of autonomous agents to interpret and interact with complex desktop interfaces. We introduce Screen2AX-Task, a benchmark specifically designed for evaluating autonomous agent task execution in macOS desktop environments. Using this benchmark, we demonstrate that Screen2AX delivers a 2.2× performance improvement over native accessibility representations and surpasses the state-of-the-art OmniParser V2 system on the ScreenSpot benchmark.
Screen2AX is open-source and available at \href{https://github.com/MacPaw/Screen2AX}{https://github.com/MacPaw/Screen2AX}.

\end{abstract}

\begin{teaserfigure}
  \includegraphics[width=\textwidth]{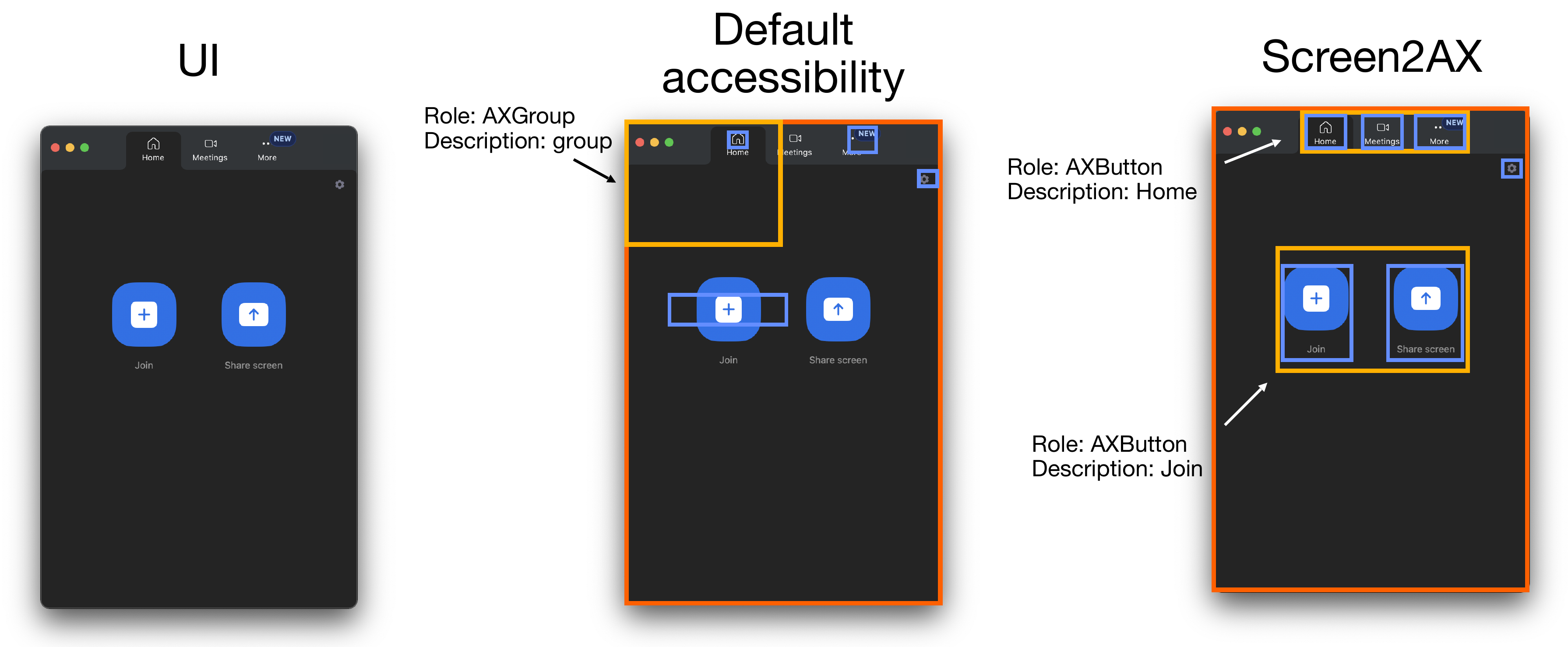}
  \caption{Screen2AX is a screenshot-based method that addresses a key gap in the availability of accessibility support by generating accessibility trees for applications that lack built-in features. Using only screenshots as input, Screen2AX expands coverage across diverse applications and can outperform built-in tools in quality.  The figure compares: (Left) the original UI screenshot, (Middle) the accessibility map from built-in tools, and (Right) the Screen2AX-generated accessibility map, demonstrating improved element recognition and hierarchical organization.}
  \Description{A side-by-side comparison of three interface screenshots. The left panel shows the original user interface of a macOS application. The middle panel displays the accessibility view generated by the built-in macOS accessibility tools, where elements are partially recognized with less structure. The right panel presents the Screen2AX-generated accessibility map, which shows detailed and correctly grouped interface elements, such as buttons labeled "Home" and "Join", organized hierarchically.}
  \label{fig:pipeline_abstract}
\end{teaserfigure}

\thanks{Preprint}

\maketitle

\section{Introduction}

Despite years of progress in accessibility standards \cite{Campbell_Adams_Montgomery_Cooper_Kirkpatrick_2024}, assistive technologies \cite{voice_over, NV_Access_2025, switch_control, universalaccess}, and platform-specific guidelines \cite{accessibility_apple, human_interface_guidelines, accessibility_inspector}, many macOS applications still fall short in providing the necessary accessibility features for users with diverse accessibility needs. Accessibility information is typically provided by the application developers, yet it is often incomplete or entirely absent.
Our preliminary analysis (detailed in Section~\ref{preliminary-analysis}) of the 99 most popular macOS applications \cite{top100} revealed that only 36\% offer structured, high-quality accessibility metadata. In contrast, 46\% include partial or low-quality metadata, and 18\% lack accessibility support entirely. A random subset of less known applications shows even more alarming results, with only 33\% providing full support, 27\% partial support, and 40\% no support at all.

This lack of comprehensive accessibility metadata directly affects the usability of desktop applications for users with disabilities. Screen readers such as VoiceOver \cite{voice_over} fully depend on provided accessibility metadata, and its incompleteness can frequently lead to misinterpretation of the position and role of user interface (UI) elements \cite{webaim_survey}, hindering effective interaction. Likewise, artificial intelligence (AI)-driven agents rely on accessibility metadata as a hierarchical representation of the screen to interpret complex UI structures. When this metadata is missing or inaccurate, these agents face significant challenges, resulting in automation failures and inconsistent performance in assistive workflows.

Desktop UIs have evolved from strictly aligned text terminals, where screen readers could read the content directly, to highly dynamic, feature-rich graphical environments. 
Today’s user interfaces incorporate windows, drag-and-drop components, and custom widgets that demand complex solutions, such as the Web Content Accessibility Guidelines and operating system (OS)-level accessibility Application Programming Interfaces (APIs), to ensure inclusive interaction \cite{Campbell_Adams_Montgomery_Cooper_Kirkpatrick_2024, accessibility_apple}. 
These frameworks introduce semantic structures, role definitions, and guidelines that enable screen readers and alternative input methods. 
However, consistent adoption across platforms remains difficult to achieve. 
In particular, desktop operating systems like macOS require developers to manually annotate or update accessibility metadata for custom UI elements, leading to incomplete coverage that affects users who depend on assistive tools\footnote{\href{https://developer.apple.com/documentation/accessibility/accessibility-api}{https://developer.apple.com/documentation/accessibility/accessibility-api}}.

In recent years, studies on accessibility generation and semantic UI understanding have been emerging, specifically in the web, iOS \cite{screen-recognition}, and Android \cite{sunkara-etal-2022-towards, Ross2018Oct} domains. Yet, to the best of our knowledge, no foundational work has addressed these challenges on macOS. We argue that this gap is largely due to the lack of comprehensive, labeled datasets for macOS, in contrast to the abundance of data available for mobile platforms \cite{VINS_dataset}.

Additionally, generating macOS accessibility metadata is more challenging due to its complexity: developers are often required to manually manage metadata for custom controls and dynamic layouts. This manual process is not only complex and time-consuming but also prone to error. Consequently, several persistent issues continue to hinder the accessibility of macOS applications, including element misclassifications, inaccurate positioning, missing element or role descriptions, and situations where elements that are not visible on the screen are still included in the metadata (as detailed in Section~\ref{accessibility-problems}).

These deficiencies simultaneously disrupt human-centric assistive tools (e.g., VoiceOver) and AI agents that rely on well-formed metadata for navigation and automation. Given the scarcity of specialized research for macOS, critical accessibility needs remain unmet.

To address these gaps, we present a vision-based system that generates macOS accessibility metadata directly using only UI screenshots. Compared to the current time-consuming manual annotation baseline, our approach employs a computer vision pipeline to detect, classify, and hierarchically group on-screen elements. We argue that such automation can ease the burden of creating complex metadata while improving application consistency. By incorporating text recognition, element detection, logical grouping, and element descriptions, our system ensures that both screen readers and AI-driven agents receive complete and accurate representations of macOS UIs.

This paper introduces three primary contributions:
\begin{itemize}
    \item \textbf{Screen2AX Framework}: an open-source\footnote{\href{https://github.com/MacPaw/Screen2AX}{https://github.com/MacPaw/Screen2AX}} deep learning framework that infers multi-level UI hierarchies and generates high-quality accessibility metadata directly from \mbox{macOS} application screenshots, using only visual input.
    \item \textbf{Screen2AX-Tree}, \textbf{Screen2AX-Element} and \textbf{Screen2AX-Group}: Three curated publicly available datasets of macOS application UIs. The Screen2AX-Tree\footnote{\href{https://huggingface.co/datasets/MacPaw/Screen2AX-Tree}{https://huggingface.co/datasets/MacPaw/Screen2AX-Tree}}  consists of screenshots paired with comprehensive, annotated accessibility structures—offering a valuable resource for future research in accessibility generation.
    The Screen2AX-Element\footnote{\href{https://huggingface.co/datasets/MacPaw/Screen2AX-Element}{https://huggingface.co/datasets/MacPaw/Screen2AX-Element}} dataset comprises detected UI elements, while the Screen2AX-Group\footnote{\href{https://huggingface.co/datasets/MacPaw/Screen2AX-Group}{https://huggingface.co/datasets/MacPaw/Screen2AX-Group}} dataset organizes these elements into meaningful groups, both providing a valuable resource for research in accessibility generation.
    \item \textbf{Screen2AX-Task\footnote{\href{https://huggingface.co/datasets/MacPaw/Screen2AX-Task}{https://huggingface.co/datasets/MacPaw/Screen2AX-Task}}}: A macOS task execution benchmark that pairs UI screenshots with corresponding task commands and target UI elements, enabling comprehensive evaluation of agent interaction and task execution.
\end{itemize}

\section{Background} \label{background}

\subsection{Accessibility Tree Structure}

Accessibility data can be represented as a hierarchical tree structure, where individual UI elements (e.g., buttons, text, images) are grouped into logical containers such as toolbars, menu bars, and other structural components. At the highest level of this tree, the full window (AXWindow) element serves as the root node that encapsulates all other UI components.
Each node of the accessibility tree has the following properties:

\begin{itemize}
    \item \textbf{Name}: a human-readable identifier for the element, often derived from the element's label or content.
    \item \textbf{Role}: the type of UI element (e.g., button, text field, image) that defines its primary function.
    \item \textbf{Description}: A textual explanation of the element’s purpose, which can provide additional context for screen readers.
    \item \textbf{Role Description}: A more user-friendly description of the element's role, often simplifying technical role names.
    \item \textbf{Value}: The current state or input value of the element, such as text in a text field or the checked state of a checkbox.
    \item \textbf{Position}: The coordinates of the element.
    \item \textbf{Size}: The width and height of the element.
    \item \textbf{Children}: A list of nested elements contained within the current node, forming the hierarchical structure of the accessibility tree.
\end{itemize}

This structure is designed to offer comprehensive coverage of the UI screen. However, in practice, a large portion of the accessibility metadata—specifically the Name, Description, Role Description, and Value fields—is frequently missing.
An example of the system accessibility tree can be found in Appendix~\ref{appendix:accessibility-tree}.

\subsection{macOS Accessibility Limitations} \label{accessibility-problems}

\begin{figure}[H]
    \centering
    \includegraphics[width=\linewidth]{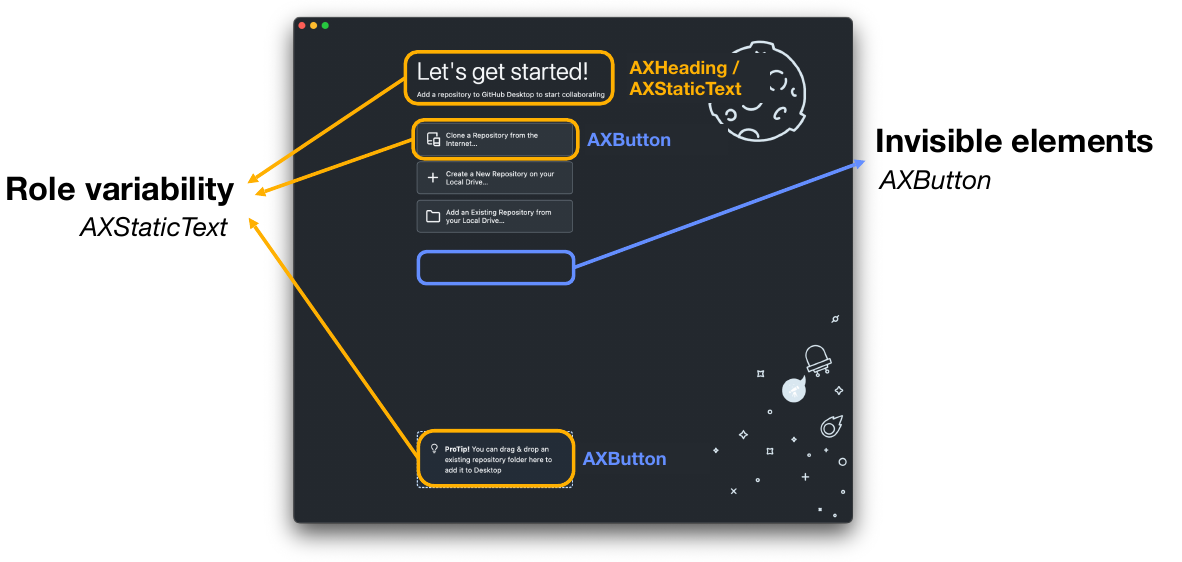}
    \caption{Limitations of built-in accessibility: Elements with inconsistent role assignments (orange) pose interpretation challenges, while visually invisible but functionally visible elements (blue) introduce ambiguity in understanding the interface.}
    \Description{A user interface screenshot of the GitHub Desktop app annotated to highlight two types of accessibility issues. Elements marked in orange have inconsistent role assignments, such as being identified as static text, while true labels must be a button or a heading/text, making interpretation unreliable. Elements outlined in blue are visually hidden but still functionally present in the accessibility tree, creating ambiguity for assistive technologies.}
    \label{fig:accessibility-limitations}    
\end{figure}

The built-in accessibility system in macOS does not always provide a reliable representation of the screen. Some UI elements may be absent from the accessibility tree, while others may contain incomplete textual descriptions or inaccurately defined bounding boxes, as illustrated in Figure~\ref{fig:accessibility-limitations}.

Although many areas need improvement in accessibility metadata, we focus on addressing the following common issues:

\begin{itemize}
    \item \textbf{Invisible elements:} Pop-ups or overlays often leave underlying UI components incorrectly tagged as visible or fail to remove elements once they disappear.
    \item \textbf{Misclassified elements:} Developers may implement a button as a generic component (e.g., a cell) and manually attach click handlers, causing screen readers to misinterpret the element’s role.
    \item \textbf{Shifted elements:} Layout glitches can displace UI elements, leading to bounding boxes that fail to match on-screen positions and confuse assistive technologies.
    \item \textbf{Missing accessibility metadata:} In some cases, elements are visible on-screen but lack accessibility metadata, making them undetectable by assistive tools. In extreme cases, the accessibility engine fails to render UI elements altogether.
\end{itemize}

\subsection{Preliminary App Analysis} \label{preliminary-analysis}

To understand the scale of the issue and explore accessibility metadata coverage, we conducted a preliminary analysis of the number of applications that have full, partial, or absent built-in macOS accessibility support.

We consider the provided accessibility tree to be \textit{full} when all elements on the screen are present in the tree, with correct corresponding bounding boxes. Additionally, smaller elements must be grouped into semantically meaningful clusters. \textit{Partial} accessibility support refers to cases where some elements are present, but some bounding boxes are incorrect, or groupings are missing. \textit{Absent} accessibility metadata means that no elements, or only three main window control elements (close, minimize, maximize), are available through the built-in accessibility.

This assessment was human-labeled and visually assessed based on plotted bounding boxes of elements and groups. Note that descriptions of elements were not included, making the calculated percentages an optimistic scenario of the accessibility coverage.

We measure the quality of accessibility on two subsets of apps:
\begin{itemize}
    \item Random Subset: Random sample of 452 macOS applications\footnote{\href{https://zenodo.org/records/15185674}{https://zenodo.org/records/15185674}} selected from Mac App Store and Brew Cask \cite{macpaw-macos-app-dataset}.
    \item Popular Apps: Top-99 free popular applications on macOS\cite{top100}.
\end{itemize}

\begin{table}[H]
  \centering
  \caption{Accessibility analysis results for macOS applications.}
  \Description{A table summarizing the accessibility levels of two groups of macOS applications: popular applications and a random subset. For popular apps, 17.7\% have no accessibility, 45.9\% have partial accessibility, and 36.5\% have full accessibility. For the random subset, 32.7\% have no accessibility, 37.8\% have partial accessibility, and 29.4\% have full accessibility.}
  \label{tab:accessibility_analysis}
  \begin{tabular}{lccc}
    \toprule
        & No metadata & Partial metadata & Full metadata \\
    \midrule
    Popular Apps & 17.7\% & 45.9\% & 36.4\% \\
    Random Subset & 32.7\% & 37.8\% & 29.4\% \\
    \bottomrule
  \end{tabular}
\end{table}

The results of our analysis show that only a third of applications provide detailed metadata support. Moreover, Table~\ref{tab:accessibility_analysis} shows a clear gap in accessibility support between popular and randomly selected macOS applications. In particular, randomly chosen apps tend to have much weaker built-in accessibility features. Overall, we find that only a small portion of macOS apps offer high-quality, well-structured accessibility metadata—highlighting the need for better tools and more consistent support for accessibility across the platform.

\section{Related Works}

Nowadays, an increasing number of people rely on various graphical user interface (GUI) assistants in their daily work. Recent advancements in assistive technologies \cite{assist-gui, show-ui, interaction-llm} have underscored the importance of effective accessibility generation for user interfaces. Accessibility refers to the hierarchical representation of a UI (see Appendix~\ref{appendix:accessibility}), which can be structured as a tree: leaf nodes represent individual UI elements (e.g., buttons, text, images), while internal nodes represent groups of elements (e.g., toolbars, menu bars).
While many traditional approaches rely on manual annotation or interaction-based data collection, recent research has explored vision-based and AI-driven techniques to automate accessibility metadata generation.

\subsection{Accessibility generation}

The process of generating accessibility metadata from an image can be broken down into several steps \cite{screen-parsing, screen-recognition}. First, UI element detection involves localizing and classifying all elements present on the screen. The next step is organizing these UI elements into logical groups (e.g., toolbar, menu bar, etc.). Finally, additional contextual information is incorporated by assigning descriptive labels to these detected groups and individual elements.

One of the first works to employ a vision-only approach for mobile accessibility generation is Screen Recognition \cite{screen-recognition}. Their pipeline for iOS consists of a UI element detection model, a heuristic-based grouping of elements, and additional models to recognize content, state, and interactivity. Specifically, they leveraged iOS's built-in Optical Character Recognition (OCR) for text detection, the icon recognition engine introduced in iOS 13, and the Image Description feature in iOS 14. However, their approach primarily clusters single elements rather than assigning semantic meaning to groups. 
In contrast, Screen Parsing \cite{screen-parsing} introduces a method for organizing UI elements into logical semantic groups, ultimately forming a structured UI graph. 

Despite recent advancements, existing methods primarily target mobile screens (iOS/Android), which are considerably simpler than desktop environments like macOS. Mobile interfaces typically feature fewer and more standardized UI elements. In contrast, our analysis shows that a typical macOS screen contains an average of 192 elements with an accessibility tree depth of 7, while Android/iOS screens contain only about 22 elements with an average depth of 3. This highlights the increased complexity of desktop interfaces, particularly in the hierarchical structure of their accessibility trees.

Element object detection in GUIs presents a challenge, as elements are often densely packed (e.g., menus), overlapping (e.g., popovers), or positioned in close proximity within a confined space. Several studies~\cite{sunkara-etal-2022-towards, ui-detection-yolo} adapted deep learning architectures (Faster R-CNN~\cite{faster-r-cnn}, CenterNet Hourglass~\cite{center-net}, DETR~\cite{detr}) for solving this task, with YOLOv8~\cite{ui-detection-yolo} showing the most promising results \cite{lu2024omniparserpurevisionbased}, offering fast inference, robust generalization, and the ability to accurately detect small elements.  

The next major step is grouping UI elements into semantically meaningful structures. One notable approach, UI Semantic Group Detection \cite{semantic-group-detection}, employs deep visual techniques—such as colormaps and prior group distribution—to group elements in mobile interfaces, achieving an F1 score of 77.5\%. However, this method focuses solely on grouping and does not address element detection or description, leaving the resulting accessibility metadata incomplete.

Among existing approaches, OmniParser V2 \cite{lu2024omniparserpurevisionbased}  is the most comparable to our work, particularly in the context of desktop UI analysis. It is a method for parsing UI screenshots into a set of structured and interpretable elements. This approach leverages a fine-tuned YOLOv8 model for UI element detection, and the Florence2 \cite{florence2} model for generating image and icon descriptions. However, while OmniParser V2 successfully extracts UI elements along with their classes and descriptions, it does not capture the hierarchical relationships between elements, which is a key aspect of our approach.

\subsection{Autonomous agents}\label{sec:autonomous-agents}

Recent advancements in Vision-Language Models (VLMs) have prompted researchers to explore their application in user interface navigation and automated task execution. In desktop environments, two primary approaches have emerged. Some studies employ a combination of screenshots and accessibility metadata to enhance the interpretability of UI elements \cite{cogagent}, whereas others have investigated vision-only methods, given the challenges associated with extracting this metadata \cite{uitars}. 

In mobile environments, several researchers have explored text-based screen representations as an alternative to vision-only approaches \cite{autodroid, mobilegpt}. However, the complexity of desktop environments, along with often insufficient accessibility metadata, makes these approaches less suitable for testing with desktop agents. Recent trends in quality assurance highlight the value of accessibility metadata, with automated test systems using large language models~(LLMs) to validate user interfaces \cite{ax-nav}. We believe that high-quality accessibility metadata is crucial grounding information for agents, enabling them to perform automation tasks more reliably.

\subsection{Data availability}

Nevertheless, the major cornerstone in generating accessibility for macOS remains data availability. There are a few publicly available datasets focused on representing UI elements and their interactions, most of which are human-annotated. For example, RICO~\cite{rico} is a large-scale dataset of Android UI screens featuring component annotations and interaction flows. Similarly, Android in the Wild \cite{aitw} captures real-world Android UI screenshots, offering a more diverse and unconstrained representation of user interfaces. 
However, macOS-specific element datasets are either not publicly available \cite{lu2024omniparserpurevisionbased} or entirely absent. Additionally, there are currently no equivalent datasets for macOS that pair UI screenshots with accessibility trees, which limits the development and benchmarking of UI accessibility models for desktop environments.

\section{Methodology}

To generate structured accessibility metadata from UI screenshots, we propose a pipeline that systematically detects UI elements, assigns descriptive labels, and organizes them into a hierarchical tree structure. This approach provides a clear and interpretable representation of the interface. The pipeline consists of the following key steps:

\begin{enumerate}
    \item \textbf{UI Element Detection}: Localize and classify all the elements displayed on the screen (e.g., buttons, images, links).
    \item \textbf{Text Detection}: Extract on-screen text using built-in Optical Character Recognition (OCR).
    \item \textbf{UI Element Description}: Generating semantic descriptions of all the elements detected in Step 1.
    \item \textbf{Grouping UI Elements}: Organize detected UI components into semantically meaningful groups (e.g., toolbars, menu bars, side panels).
    \item Build a \textbf{hierarchical representation} of the UI, capturing parent-child relationships and assigning the groups to the appropriate hierarchical levels (e.g., as siblings or nested structures) to accurately reflect the overall interface layout.
\end{enumerate}

\begin{figure*}
    \centering
    \includegraphics[width=\linewidth]{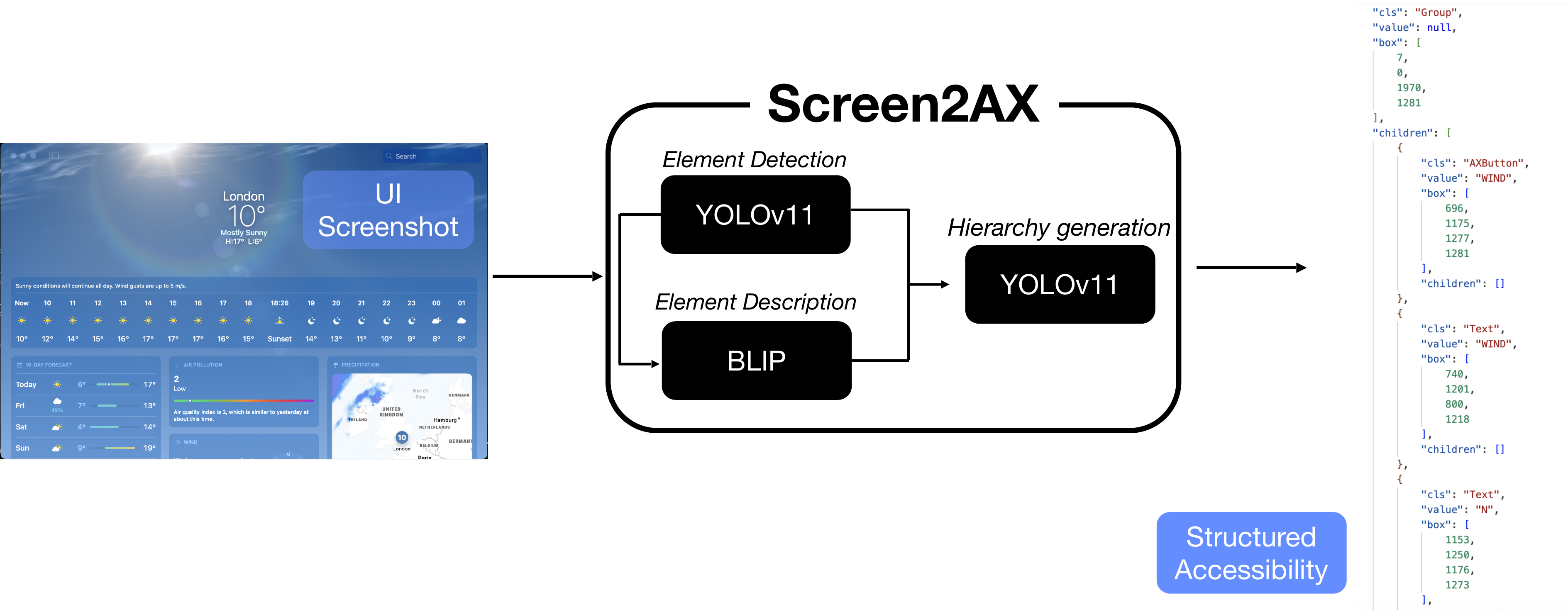}
    \caption{Screen2AX pipeline for automated accessibility generation. The system processes a UI screenshot through three stages: (1) YOLOv11 detects UI elements, (2) BLIP generates element descriptions, and (3) a second YOLOv11 model organizes elements along with their descriptions into a hierarchical structure. The final output is structured hierarchical accessibility metadata.}
    \Description{A visual overview of the Screen2AX pipeline. Starting from a UI screenshot (left), the pipeline processes the image in three main stages. First, a YOLOv11-based object detection module identifies UI elements. Next, each detected element is passed to a BLIP-based model to generate textual descriptions. Finally, a second YOLOv11 model infers the layout structure to create a hierarchical organization of elements. The result is structured accessibility metadata that captures both content and spatial relationships, enabling downstream accessibility services.}
    \label{fig:pipeline-methodology}
\end{figure*}

An overview of the proposed pipeline is illustrated in Figure~\ref{fig:pipeline-methodology}.

\subsection{Datasets}\label{sec:datasets}

Generally, the built-in macOS accessibility classifies the UI elements across 52 classes (the full list can be seen in Appendix ~\ref{appendix:accessibility}). While class information in accessibility is informative and can provide extra value, most of the classes are rarely observed, so our research focuses on a simplified categorization consisting of seven key classes. The chosen classes are:
\begin{itemize}
    \item \textbf{AXButton}: This category encompasses all interactable button-like elements, including AXCheckBox and AXRadioButton, which are not always clearly distinguishable in practice and are thus merged.
    \item \textbf{AXDisclosureTriangle}: This element shows and hides information and is functionality associated with a view or a list of items. Although technically a subtype of a button, it has a frequent occurrence and a specific function. 
    \item \textbf{AXLink}: Represents clickable text elements that perform an action upon interaction.
    \item \textbf{AXTextArea}: Consolidates all input-related elements, such as text fields and editable areas.
    \item \textbf{AXImage}: Groups together both decorative and functional images, including icons.
    \item \textbf{AXStaticText}: All text field, including headings. 
    \item \textbf{AXGroup}: All element groupings that provide structural information, including menus, logical sections, or top bars.
\end{itemize}

To build datasets for accessibility metadata generation, we collected 1127 screenshots from 112 macOS applications, along with their built-in accessibility metadata, using the open-source \textit{macapptree} library\footnote{\href{https://github.com/MacPaw/macapptree}{https://github.com/MacPaw/macapptree}}. Since accessibility metadata is often incomplete or inaccurate, we manually corrected the system labels using Roboflow platform\footnote{\href{https://roboflow.com}{https://roboflow.com}} to fix existing annotations. The annotations were created for dedicated datasets, including UI element detection, element grouping, and element selection screen-level tasks for assessing AI agents.

\subsubsection{Screen2AX-Tree: System Accessibility Metadata}
\begin{figure}[H]
    \centering
    \begin{minipage}{\linewidth}
        \centering
        \includegraphics[width=\linewidth]{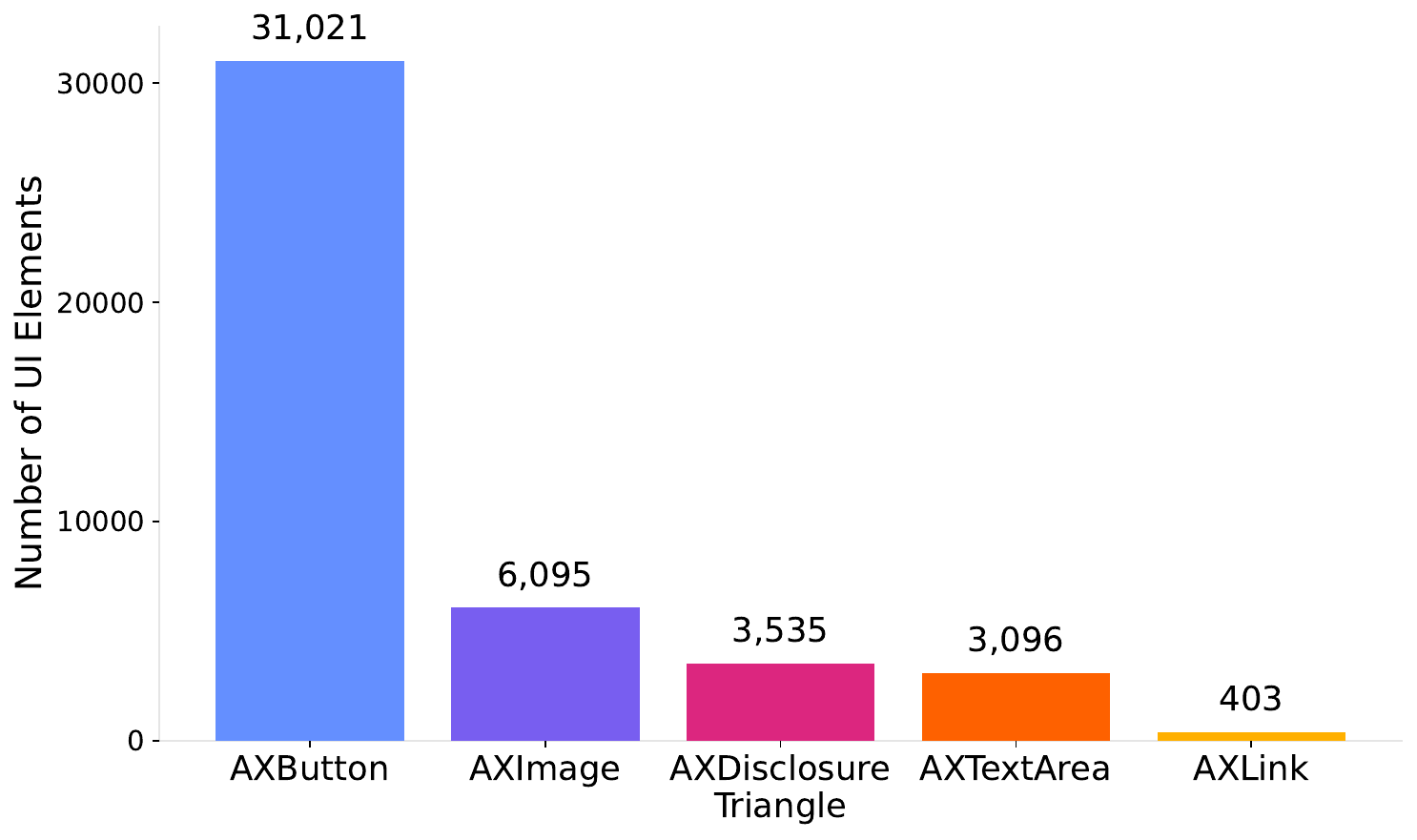}
        \caption{Class distribution of UI elements in the Screen2AX-Element dataset. The figure highlights a significant class imbalance, with AXButton dominating the other classes.}
        \Description{Class distribution of elements in the dataset. The bar plot shows that there are 29627 elements of type AXButton, 3898 AXImage, 3545 AXDisclosureTriangle, 3171 AXTextArea, and 400 AXLink. This suggests that the dataset is highly imbalanced.}
        \label{fig:UI_elements}
    \end{minipage}\hfill
\end{figure}

Screen2AX-Tree comprises UI screenshots and raw system accessibility metadata from 112 macOS applications, resulting in 1127 images with annotations spanning 52 UI element classes (listed in Appendix~\ref{appendix:accessibility}). Additionally, we verified that the proportion of light and dark theme screenshots in our dataset is 52\% and 48\%, respectively, indicating a well-balanced distribution. After filtering and merging relevant classes, we obtained a Screen2AX-Element dataset. Figure~\ref{fig:UI_elements} illustrates the distribution of elements within the dataset.

\subsubsection{Screen2AX-Element: UI Element Detection}
Screen2AX-Element is a dataset targeting UI element detection problems. It has 986 images with a total of 44,150 annotations across five key classes: \textit{AXButton, AXDisclosureTriangle, AXImage, AXLink}, and \textit{AXTextArea}. While experimenting with element detection, we found that AXStaticText is best detected using OCR, so it's not included in this dataset.

\begin{figure}[H]
    \centering
    \includegraphics[width=\linewidth]{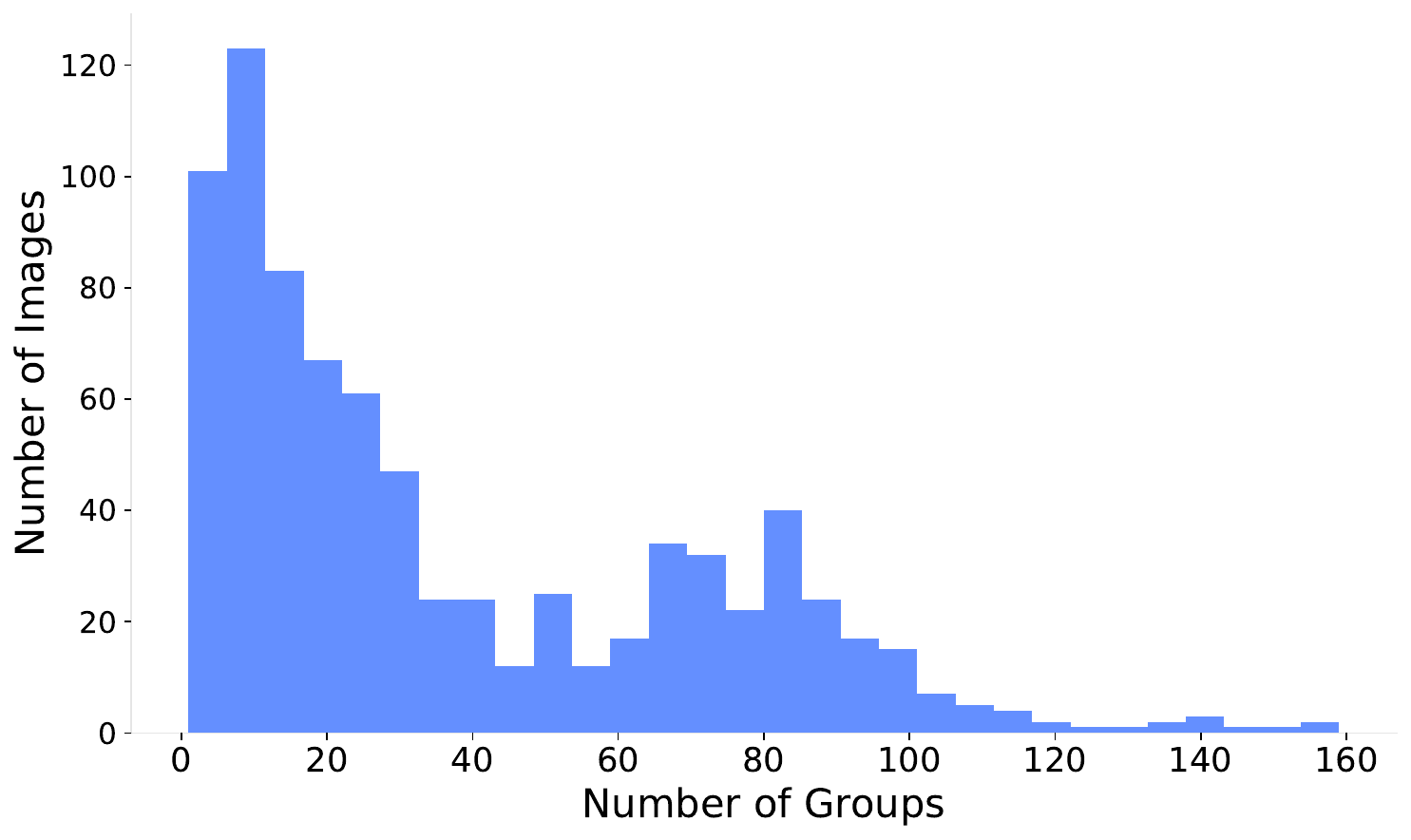}
    \caption{Histogram showing the bimodal distribution of groups per image in Screen2AX-Group.}
    \Description{A histogram showing the distribution of UI groups per image in  Screen2AX-Group. The X-axis represents the number of groups, ranging approximately from 0 to 160. The Y-axis represents the number of images. There are two prominent peaks in the histogram—one around 20 groups and another around 80 groups per image.}

    \label{fig:groups-hist}
\end{figure}  

\subsubsection{Screen2AX-Group: UI Hierarchy Prediction}
The Screen2AX-Group dataset comprises hierarchical relationships between elements, and, similarly to Screen2AX-Element, it is represented as bounding boxes. We filtered 808 images with 30,774 annotations across all the images. The resulting dataset includes 19 types of groups, but during the later model training for group detection, we merged all classes into a single class \textit{AXGroup}.

As shown in Figure~\ref{fig:groups-hist}, the distribution of groups per image is bimodal. The first and largest peak occurs around 20 groups per application, representing interfaces with a minimal number of groups—either simple applications or ones where developers omitted explicit structuring. The second peak appears around 80 groups per application, reflecting highly segmented or complex interfaces.

\subsubsection{Screen2AX-Task: UI Screen Task Evaluation}
Screen2AX-Task dataset was gathered to evaluate how the quality of accessibility metadata impacts the ability of AI-based agents, as discussed in Section~\ref{sec:autonomous-agents}. To support this use case, we comprised 435 images, selected as a representative subsample from the screenshots collected for the Screen2AX-Tree. Notably, evaluation is done on a subset of only 166 images—none of which were used during training or validation of the models at any step.

Each image in the dataset includes a box of interactive elements annotated by a human and a caption generated on top of it using GPT-4 \cite{openai2023gpt4}. The dataset consists of the following fields: \textit{x1}, \textit{y1}, \textit{x2}, \textit{y2}, \textit{image\_width}, \textit{image\_height},  \textit{command} and  \textit{visual description}. The coordinates \textit{x1}, \textit{y1}, \textit{x2}, and \textit{y2} define the bounding box of a UI element within the screenshot. The fields \textit{image\_width} and \textit{image\_height} define the dimensions of the image. The \textit{command} field specifies the intended user action (e.g., "click the button", "open menu") - desired element functionality, while the \textit{visual description} provides a textual description of the UI element (e.g., "red minimize button", "submit form button"). In total, this dataset has 5934 rows of annotated commands and visual descriptions and is used to evaluate our approach later.

The dataset contains a diverse range of UI elements, including icons (55.3\%), buttons (19.7\%), checkboxes (8.1\%), and others, such as links, dropdowns, and menus. Commands are created using simple and direct language. Most commands (68.8\%) are of moderate complexity with 3-4 words,  with the most complex ones containing up to 10 words (e.g., "select the option to move windows with minimum sizes offscreen"). The dataset exhibits significant variation in element sizes and spatial distribution across interfaces (see Appendix~\ref{appendix:task-analysis}). This dataset serves as a robust benchmark for testing AI agents' ability to understand screen content, interpret simple one-step instructions, and correctly identify the appropriate UI elements for interaction.

\subsection{UI element detection}
Detecting UI elements is the foundational step in our accessibility generation pipeline. Element detection consists of 2 smaller subtasks: localization of elements in the image and their classification. The goal is to accurately find and classify all visible UI components on a macOS screen—such as buttons, text fields, images, and links. This step produces a set of detected elements, each categorized into one of five predefined classes (\textit{AXButton, AXDisclosureTriangle, AXImage, AXLink, AXTextArea}) and annotated with their corresponding bounding box coordinates, based on the Screen2AX-Element dataset.

\subsubsection{Deterministic UI Element Detection: MSER + OCR}

Our first approach, used as a baseline, combines the Maximally Stable Extremal Regions (MSER) algorithm \cite{MATAS2004761} with Optical Character Recognition (OCR) in a deterministic method.

Initially, we detect all text in the image using the ocrmac\footnote{\href{https://github.com/straussmaximilian/ocrmac}{https://github.com/straussmaximilian/ocrmac}} tool. Afterward, the detected text is removed to prevent interference with the MSER algorithm. We then apply the MSER algorithm to the image, filter and merge the resulting bounding boxes, and identify the final set of UI element boxes. The core principle of MSER is to identify stable, distinctive regions, which, in this context, correspond to UI elements.

Once UI elements are detected, the next step is classifying them into predefined categories. We explored two approaches for this classification task:

\begin{enumerate}
    \item GPT-4-Based Classification: Each detected element was overlaid with a green bounding box and assigned a unique identifier. We then prompted GPT-4 to predict the class of each element or to indicate any segmentation errors. The detailed prompt can be found in Appendix~\ref{appendix:classification-prompt}.
    \item YOLOv11 Model for Classification: We trained a YOLOv11-based model on the Screen2AX-Element dataset to classify detected UI elements into five categories.
\end{enumerate}

However, these approaches did not yield satisfactory results (see Section~\ref{sec:results}). Consequently, we developed an alternative deep-learning-based solution to improve classification accuracy.

\subsubsection{YOLOv11 for UI Element Detection}

YOLOv11\footnote{\href{https://github.com/ultralytics/ultralytics}{https://github.com/ultralytics/ultralytics}} has demonstrated strong performance in UI element detection (i.e. localization and then classification), offering fast inference, robust generalization, and the ability to accurately detect small elements. Additionally, prior work \cite{ui-detection-yolo} has shown that YOLOv8 achieves high precision and recall (over 90\%) in similar UI detection tasks, further supporting its suitability for our approach. Therefore, we fine-tuned the YOLOv11 model on Screen2AX-Element for element detection.

Before training, we resized all images to ensure consistency across samples. YOLOv11 also incorporates internal mechanisms for data augmentation and specializes in training loss functions to mitigate class imbalance, which is particularly relevant in our case.
Additionally, we filtered out elements with low confidence scores to reduce false positives and improve detection reliability.

\subsection{UI element description}

The coordinates and class of UI elements alone are not sufficient to fully understand their functionality. This is especially important for buttons, as we need to determine their intended actions when clicked. 

Unlike mobile interfaces, which mostly use text alongside icons, desktop applications rely more heavily on icon-only buttons, where the graphical representation of the button is key to conveying its functionality. Our analysis revealed that 75\% of the buttons in our training dataset contain text labels. If a button has text, we use it as its description, as the text provides the most explicit characterization of the button’s function. However, if a button lacks text, we rely on a model to generate a descriptive label based on its icon.

\subsubsection{Text detection}

Since our research focuses on macOS applications, we selected the ocrmac tool, which leverages Apple’s Vision framework. This tool demonstrated excellent performance for our task, offering both high speed and accuracy compared to other available OCR approaches.

\subsubsection{BLIP for element description}

If a button does not contain text, we need to generate a caption using only an image of a button as input. Therefore, we fine-tuned the Bootstrapping Language-Image Pre-training(BLIP) model \cite{blip} for icon captioning.

For fine-tuning, we used a publicly available dataset \cite{haque2024inferringalttextuiicons}, which consists primarily of mobile UI icons. We fine-tuned the model on 5,000 icons from the dataset over 10 epochs using the AdamW\cite{loshchilov2019decoupledweightdecayregularization} optimizer with a learning rate of 5e-5. Despite the dataset being designed for mobile platforms, we consider it suitable for our task since many UI icons are visually similar across platforms. Additionally, we collected a dataset of 988 macOS icons and captioned it using GPT-4 \cite{openai2023gpt4} for validation and testing purposes.

\begin{figure*}[htbp]
    \centering
    \includegraphics[width=0.9\linewidth]{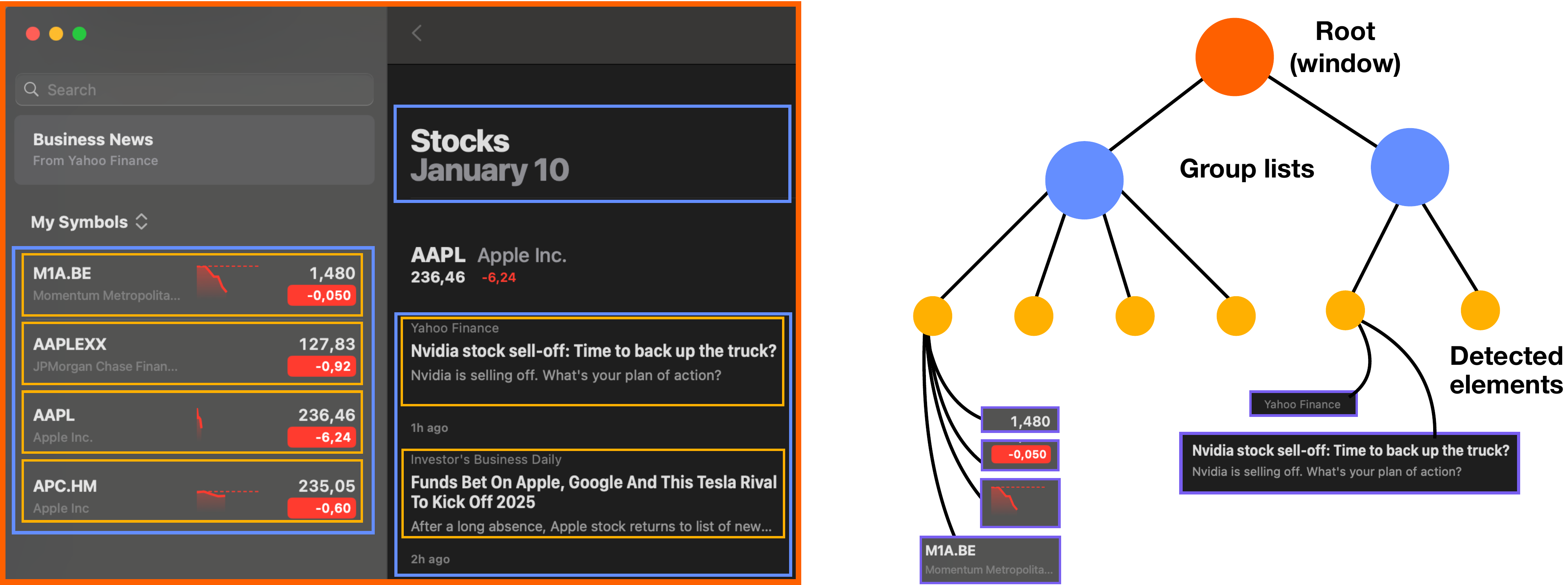}
    \Description{Example of hierarchy tree}
    \caption{An example of the labeled input screen (left) and the corresponding hierarchy tree (right).
    The tree captures every element in the screenshot, with leaf nodes for individual UI elements (purple), middle group nodes for grouped elements (yellow), and higher hierarchy nodes for logical sections (blue). The root node (orange) represents the entire window. Connecting lines highlight the relationship between interface elements and their hierarchical structure.}
    \Description{The figure shows a labeled UI screen on the left and its corresponding hierarchy tree on the right. Each UI element on the screen is represented as a node in the tree. Leaf nodes are colored purple and represent individual UI components. Yellow nodes group multiple related elements together. Blue nodes indicate higher-level logical sections. An orange root node at the top of the tree represents the entire window. Lines connect nodes to illustrate parent-child relationships in the UI structure.}
    \label{fig:tree_hierarchy}
\end{figure*}

\subsection{Hierarchy Generation}
Structuring UI elements into logical groups to form a hierarchical representation is a critical step in generating high-quality accessibility metadata.
This process organizes the interface into a tree structure, where leaf nodes represent individual UI elements and higher-level nodes encapsulate semantically meaningful groups. 
A well-formed hierarchy provides assistive technologies with a clear and organized view of on-screen content, enhancing both screen reader navigation and AI-driven interactions. An example of such a structured hierarchy is shown in Figure~\ref{fig:tree_hierarchy}.

We employ two distinct methods for hierarchy generation:

\begin{enumerate}
    \item \textbf{A heuristic-based approach}, which relies on spatial relationships and element properties to form groups deterministically.
    \item \textbf{A deep learning-based approach}, which utilizes a pre-trained model to detect groups.
\end{enumerate}

\subsubsection{Heuristic-Based Approach}
As a baseline, we adopt a heuristic-based method similar to the grouping mechanism employed in Screen Recognition \cite{screen-recognition}. This approach leverages element positioning, visual characteristics, and interaction-based patterns to define UI groups. The core heuristics include:

\begin{itemize}
    \item \textbf{Text Grouping:} 2 consecutive text elements $t_1$ and $t_2$ are grouped if they satisfy the following conditions:
    \begin{enumerate}
        \item Their bounding boxes overlap along the x-axis.
        \item Their vertical spacing is below a threshold, set as 
        
        $\min[\text{height}(t_1), \text{height}(t_2)] + 15$ pixels.
    \end{enumerate}
    
    \item \textbf{Image and Caption Association:} We associate images or buttons with adjacent captions based on spatial proximity:
    \begin{enumerate}
        \item The elements overlap by at least 25\% along the x-axis, with vertical spacing below 2\% of the screen height.
        \item The elements overlap by at least 40\% along the y-axis, with horizontal spacing below 2\% of the screen width.
    \end{enumerate}
    
    \item \textbf{Column Formation:} 2 elements $b_1$ and $b_2$ are considered part of the same column if they meet both of the following criteria:
    \begin{enumerate}
        \item Their vertical spacing $< 1.25 \cdot \min[\text{height($b_1$)}, \text{height($b_2$)}]$.
        \item The difference between their left or right edges is less than 40 pixels.
    \end{enumerate}

    \item \textbf{Color-Based Grouping:} Color similarity is often indicative of structural relationships within UI components:
    \begin{enumerate}
        \item Detect the most frequent colors on the screenshot.
        \item Generate a binary mask isolating pixels of a given color.
        \item Apply Morphological Opening to remove noise.
        \item Extract bounding boxes of color regions as candidate groups.
    \end{enumerate}
    
    \item \textbf{Row Formation:} Similar to column detection, row-based grouping ensures horizontal alignment is captured appropriately.
\end{itemize}

While the heuristic-based approach offers a deterministic means of structuring UI elements, it introduces several limitations. For example, the method relies on 15 hyperparameters, requiring manual fine-tuning per application to achieve optimal results. The sensitivity of these thresholds reduces the method's robustness when applied to diverse UI layouts, leading to inconsistencies in hierarchy formation.

\subsubsection{Deep Learning-Based Approach}
To address the limitations of heuristic-based grouping, we introduce a deep learning-based alternative that leverages an object detection model to identify UI groups. Specifically, we fine-tune the YOLOv11 architecture for UI group detection, defining a single class: \textit{AXGroup}. This model learns to recognize semantically meaningful clusters of elements based on a dataset that was introduced earlier.

By integrating deep learning, we ensure a scalable and adaptable solution for accessibility hierarchy generation, significantly reducing reliance on predefined heuristics while improving the accuracy of UI grouping.

\section{Results \& Evaluation}
\label{sec:results}
In this section, we present the evaluation methodology and results for each step of the Screen2AX framework. We assess the performance of UI element detection, UI element description, and hierarchy generation using quantitative metrics and qualitative analysis. 

\subsection{UI Element Detection}

To evaluate the effectiveness of the UI Element Detection step, we conducted a comparative analysis of different approaches. Specifically, we examined a deterministic method based on the Maximally Stable Extremal Regions (MSER) algorithm and a deep learning-based approach leveraging a YOLOv11 model for object detection. Our evaluation focused on the accuracy of predicted bounding boxes and classification performance across different methods.

The localization accuracy is defined as the ratio of correctly predicted boxes to the total number of boxes. To calculate localization accuracy, we measured the proportion of correctly predicted bounding boxes, considering a prediction correct if it exhibited an Intersection over Union (IoU) greater than 0.5 with the ground truth bounding box.

\begin{table}[H]
  \centering
  \caption{Comparison of object detection accuracy.}
  \Description{A table comparing object detection accuracy between two methods: a deterministic approach and YOLOv11. The accuracy at the 50\% threshold is 37.8\% for the deterministic method and 65.4\% for YOLOv11, indicating that YOLOv11 performs significantly better.}
    
  \label{tab:object_detection_metrics}
  \begin{tabular}{lcc}
    \toprule
    & Accuracy@50, \% \\
     \midrule
    Deterministic & 37.8 \\ 
    YOLOv11 & \textbf{65.4} \\
    \bottomrule
  \end{tabular}
\end{table}

The results in Table~\ref{tab:object_detection_metrics} demonstrate that the YOLOv11-based method significantly outperforms the MSER-based approach, achieving an accuracy of 65.4\% compared to 37.8\%.

After box localization, the next step involved classifying the identified UI elements. We evaluated three different classification approaches:

\begin{itemize}
    \item MSER + GPT-4: Using MSER for localization and GPT-4 for classification.
    \item MSER + YOLOv11: Using MSER for localization and YOLOv11 for classification.
    \item All YOLOv11: Using YOLOv11 for element detection (which includes localization and classification).
\end{itemize}

To assess classification performance, we employed standard evaluation metrics, including Precision, Recall, F1 score, mean Average Precision at IoU 50 (mAP50), and mean Average Precision at IoU 50-95 (mAP50-95). Additionally, since computational efficiency is crucial for real-time accessibility generation, we measured the execution time of each method on a MacBook M1 Pro.

\begin{table}[H]
  \centering
  \caption{Classification performance comparison.}
  \Description{A table comparing the classification performance of three methods: MSER+YOLOv11, MSER+GPT-4, and All YOLOv11. The All YOLOv11 approach outperforms the others across all metrics. It achieves the highest precision at 49.1\%, recall at 43.4\%, F1 score at 43.8\%, mAP@50 at 46.6\%, and mAP@50-95 at 32.0\%. It is also the fastest, with a processing time of 0.2 seconds. In comparison, MSER+YOLOv11 and MSER+GPT-4 show significantly lower performance across all metrics.}
    
  \label{tab:classification_metrics}
  \begin{tabular}{lccc}
    \toprule
    & MSER+YOLOv11 & MSER+GPT-4 & All YOLOv11 \\
    \midrule
    Precision, \% & 8.5 & 14.1 & \textbf{49.1} \\
    Recall, \% & 5.2 & 5.9 & \textbf{43.4} \\
    F1, \% & 6.5 & 8.1 & \textbf{43.8} \\
    mAP50, \% & 2.7 & 5.4 & \textbf{46.6} \\
    mAP50-95, \% & 1.5 & 2.8 & \textbf{32.0} \\
    Time, s & 1.237 & 13.28 & \textbf{0.204} \\
    \bottomrule
  \end{tabular}
\end{table}

As shown in Table~\ref{tab:classification_metrics}, the "All YOLOv11" approach consistently outperforms other methods across all evaluation metrics. Notably, it achieves an F1 score of 43.8\%, significantly higher than MSER+GPT-4 (8.1\%) and MSER+YOLOv11 (6.5\%). Additionally, the "All YOLOv11" method is the fastest, with an average execution time of just 0.204 seconds, making it the most practical choice for real-time accessibility metadata generation.

Table~\ref{tab:full_width_metrics} provides an in-depth analysis of performance across different UI element categories. The results reveal that MSER-based methods struggle with categories such as AXDisclosureTriangle, AXLink, and AXTextArea—likely due to their small size and low contrast. Moreover, the confusion matrices shown in Figures~\ref{fig:deterministic_cm}~and~\ref{fig:yolo_cm} (see Appendix~\ref{appendix:confusion-matrices} for further details) highlight the classification errors associated with both the MSER+YOLOv11 and YOLOv11 approaches. These findings underscore the significant advantage of the YOLOv11-based model in accurately classifying UI elements.

\begin{table*}[ht]
  \centering
  \caption{Per-category classification performance.}
  \Description{A table presenting per-category classification performance across five accessibility element types: AXButton, AXDisclosureTriangle, AXImage, AXLink, and AXTextArea. Three methods are compared: MSER+YOLOv11, MSER+GPT-4, and All YOLOv11. The All YOLOv11 method consistently achieves the highest scores across all metrics and categories. MSER+YOLOv11 and MSER+GPT-4 show much lower values, with zero scores in several categories, such as AXDisclosureTriangle and AXLink. This highlights the superiority of the All YOLOv11 approach for fine-grained accessibility classification.}

  \label{tab:full_width_metrics}
  \begin{tabular}{lccccc}
    \toprule
        & AXButton & AXDisclosureTriangle & AXImage & AXLink & AXTextArea \\
    \midrule
    \multicolumn{6}{c}{\textbf{Precision, \%}} \\
    \midrule
    MSER+YOLOv11 & 8.3  & 0  & 16.9  & 0  & 0  \\
    MSER+GPT-4 & 12.3   & 0  & 31.5  & 0  & 0.7  \\
    All YOLOv11  & \textbf{48.6} & \textbf{71.4} & \textbf{57.4} & \textbf{78.8} & \textbf{36.3} \\
    \midrule

    \multicolumn{6}{c}{\textbf{Recall, \%}} \\
    \midrule
    MSER+YOLOv11 & 5.5  & 0  & 9.2  & 0  & 0  \\
    MSER+GPT-4 & 3.7   & 0  & 17.6  & 0  & 0.3  \\
    All YOLOv11  & \textbf{57.5} & \textbf{21.3} & \textbf{24.1} & \textbf{38.2} & \textbf{16.2} \\
    \midrule

    \multicolumn{6}{c}{\textbf{F1 Score, \%}} \\
    \midrule
    MSER+YOLO & 6.6  & 0  & 11.9  & 0  & 0  \\
    MSER+GPT-4 & 5.6  & 0  & 22.5  & 0  & 0.4  \\
    All YOLO  & \textbf{52.7} & \textbf{32.8} & \textbf{33.9} & \textbf{51.5} & \textbf{22.4} \\
    \midrule

    \multicolumn{6}{c}{\textbf{AP50, \%}} \\
    \midrule
    MSER+YOLOv11 & 4.5  & 0  & 9.3  & 0  & 0  \\
    MSER+GPT-4 & 6.8  & 0  & 19.9 & 0  & 0.4  \\
    All YOLOv11  & \textbf{57.2} & \textbf{46.8} & \textbf{41.2} & \textbf{60.6} & \textbf{27.5} \\
    \bottomrule
  \end{tabular}
\end{table*}

\subsection{UI Element Description}

To evaluate the effectiveness of our BLIP model for element description, we use standard metrics, including and CIDEr \cite{cider} score, along with GPT-based accuracy assessment. 

CIDEr \cite{cider} score is a metric based on BLEU \cite{bleu} but is specifically designed for measuring the quality of image captioning tasks. In our case, we consider it to be from 0 to 1, with 1 indicating perfect similarity between two sentences.

The GPT-measured accuracy is calculated by providing GPT-4\cite{openai2023gpt4} with the ground-truth caption, the predicted caption, and the corresponding image. GPT-4 is then asked to determine whether the ground truth and predicted captions have the same meaning, assigning a label of 1 if they do and 0 otherwise. The prompt is available in Appendix~\ref{appendix:gpt-accuracy-prompt}. The accuracy is then calculated as the ratio of the number of correctly predicted captions to the total number.
See the results of the evaluation in Table ~\ref{tab:icon_metrics}.

\begin{table}[H]
  \centering
  \caption{Comparison of icon captioning performance.}
  \Description{A table showing the performance of a BLIP-tuned model for icon captioning. The model achieves a CIDEr score of 0.68 and a GPT-measured accuracy of 0.76, indicating strong performance in generating relevant and accurate captions.}

  \label{tab:icon_metrics}
  \begin{tabular}{lcc}
    \toprule
        & CIDEr & Accuracy (GPT-measured) \\
    \midrule
    Florence (OmniParser V2) & 0.39 & 0.44 \\
    BLIP (Screen2AX) & 0.68  & 0.76   \\
    \bottomrule
  \end{tabular}
\end{table}

Overall, the table suggests that the descriptions are generally accurate; however, the model struggles with examples that differ significantly from typical mobile icons.

\subsection{Hierarchy Generation}
To evaluate the accuracy of hierarchy generation, we adopt the same set of metrics as used in prior work \cite{screen-parsing}. Specifically, we employ three key metrics: Edge-based, Distance-based, and Group-based metrics.

\subsubsection{Edge-based Metrics}
This metric assesses hierarchy generation by decomposing the resulting tree into a set of edges. The evaluation is performed by computing the F1 score for both edge detection and leaf detection. The metric is bounded between 0 and 1, where higher values indicate better performance. A score of 0 signifies that no corresponding edges were identified in the predicted hierarchy.

\subsubsection{Distance-based Metric}
To quantify the structural similarity between the generated hierarchy and the ground truth, we utilize Graph Edit Distance (GED). GED measures the minimum number of elementary operations required to transform one graph into another. The GED calculation is defined as follows:

$$
    \displaystyle GED(g_{1},g_{2})=\min _{(e_{1},...,e_{k})\in {\mathcal {P}}(g_{1},g_{2})}\sum _{i=1}^{k}c(e_{i}) 
$$

Since computing GED exactly is NP-complete and computationally expensive, we employ an optimized approach using the \textit{optimize\_graph\_edit\_distance} method from the NetworkX library \cite{SciPyProceedings_11}. This method provides an upper-bound estimate of GED, and we use the first computed value from the iterator. If the computation exceeds 10 seconds, we approximate the GED as the number of edges in the ground truth tree. Unlike the previous metric, GED is unbounded, ranging from 0 to infinity, where lower values indicate a closer match to the ground truth hierarchy.

\subsubsection{Group-based Metric}
While the previous metrics assess the accuracy of hierarchical structure, the Group-based metric evaluates the correctness of group detection. This is particularly useful when multiple valid hierarchies exist for a given interface. The metric is computed as the mean Intersection over Union (IoU) between intermediate nodes in the generated and ground truth hierarchies. Similarly to the Edge-based metric, the Group-based metric is bounded between 0 and 1, where higher values indicate better group detection performance. This metric was previously referred to as the Container Match (CM) score in prior research \cite{screen-parsing}.

In addition to accuracy metrics, we evaluate the computational time required for hierarchy generation using different methods, excluding the UI Elements Description step. For UI element detection, we utilize the YOLOv11 model due to its superior performance and compare its effectiveness for group detection against the earlier described heuristic-based method.

It is also important to note that our approaches were evaluated on our dataset, which serves as the ground truth. However, since the dataset was collected from real-world applications with manually generated accessibility data, it may contain inaccuracies or inconsistencies, as described in Section~\ref{accessibility-problems}.

\begin{table}[ht]
  \centering
  \caption{Comparison of hierarchy generation methods.}
    \Description{A table comparing the performance of two hierarchy generation methods: Heuristics and YOLOv11. YOLOv11 outperforms Heuristics across most metrics. It achieves a higher CM score of 55\% compared to 24\% and a better overall F1 score of 77\% versus 67\%. The Leaves F1 score is also slightly higher for YOLOv11 at 31\% compared to 25\%. Additionally, YOLOv11 achieves a lower Graph Edit Distance (GED), indicating a more accurate hierarchy structure, with 200.28 versus 208.71. However, Heuristics is faster, with a generation time of 0.61 seconds compared to 0.76 seconds for YOLOv11.}
  \label{tab:hierarchy_comparison}
  \begin{tabular}{lcc}
    \toprule
    & Heuristics & YOLOv11 \\
    \midrule
    CM, \% & $24$ & \textbf{55} \\
    F1 Score, \% & $67$ & \textbf{77} \\
    Leaves F1 Score, \% & $25$ & \textbf{31} \\
    GED & $218.71$ & \textbf{200.28} \\
    Generation Time, s & \textbf{0.61} & $0.76$ \\
    \bottomrule
  \end{tabular}
\end{table}

The results presented in Table~\ref{tab:hierarchy_comparison} highlight the strengths and weaknesses of the two hierarchy generation approaches. YOLOv11 outperforms the heuristic method in almost all metrics, indicating a more accurate grouping of UI elements. Notably, YOLOv11 achieves a lower Graph Edit Distance (GED), indicating a closer match to the ground truth hierarchy while requiring slightly more computation time. 

\subsection{AI Agents Performance}

In this subsection, we evaluate how accessibility metadata from Screen2AX can help AI agents solve UI-related tasks on two benchmarks: ScreenSpot \cite{screen-spot} and Screen2AX-Task presented in Section~\ref{sec:datasets}.

\begin{figure*}[htbp]
    \centering
    \includegraphics[width=0.95\linewidth]{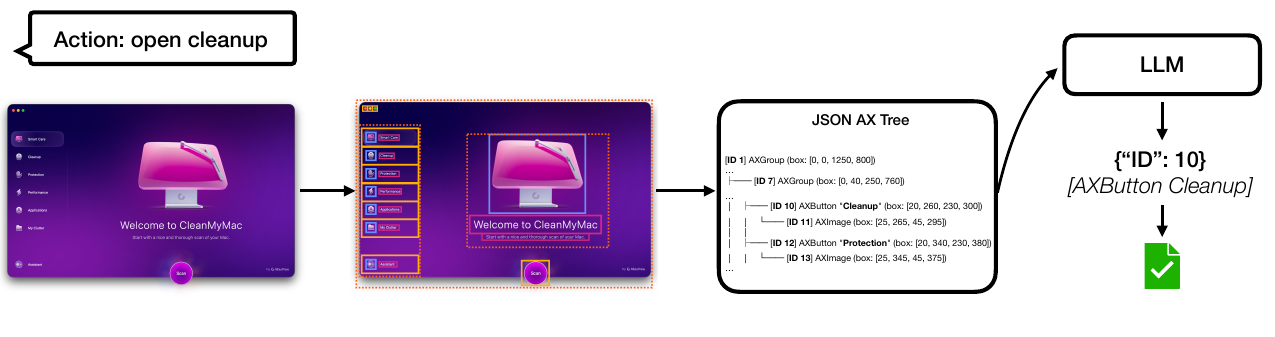}
    \caption{AI Agent Task Execution Pipeline Using Screen2AX Metadata}
    \Description{An overview of how accessibility metadata generated by Screen2AX enables a language model (GPT-4) to perform UI interaction tasks. The pipeline begins with a UI screenshot where the user's task is to "open cleanup." Screen2AX detects and groups UI elements and then converts them into a structured JSON-based accessibility (\textit{AX}) tree. Each element in the tree is assigned a unique ID, along with its role, label, and bounding box. The JSON tree is passed to an LLM (in this case, GPT-4), which is prompted with both the accessibility metadata and the target action. The model successfully maps the task to the correct element ID—in this example, \textit{{ "ID": 10 }} corresponding to the "Cleanup" button—and returns the result for execution, demonstrating how Screen2AX improves screen understanding and interaction for autonomous agents.}
    \label{fig:agent}
\end{figure*}

Accessibility metadata can empower AI agents to navigate screens and interact with applications to perform various tasks \cite{gui-agent, assist-gui}. Accordingly, we assess our method's effectiveness by evaluating the performance of an AI agent. In our experiments, we employ GPT-4 as the underlying language model to interpret the accessibility data. To simplify processing, we assign a unique identifier to each accessibility element, numbering them sequentially from $1$ to $n$, where $n$ is the total number of elements in the metadata.
The pipeline of the task performance is shown in Figure~\ref{fig:agent}.

\subsubsection{ScreenSpot evaluation}

Firstly, we evaluate our method on the publicly available ScreenSpot benchmark~\cite{screen-spot}. This benchmark is designed to assess the ability of Vision-Language Models (VLMs) to locate screen elements based on provided instructions in commonly used applications.

For our evaluation, we selected only screenshots from macOS system applications, yielding 172 task instructions. We measured performance using the Success Rate, defined as the ratio of samples in which the center point of the predicted bounding box falls within the corresponding ground truth bounding box to the total number of samples. This metric effectively quantifies the frequency with which the model's predictions result in successful UI element selection.

We use GPT-4 \cite{openai2023gpt4} as the vision-language model (VLM) in our experiments. We evaluate four different input representations: (1) image-only, and three accessibility-focused representations without images—(2) OmniParserV2 data, (3) Screen2AX without hierarchy, and (4) Screen2AX.

\begin{table}[H]
  \centering
  \caption{ScreenSpot evaluation.}
  \Description{A table showing the click accuracy of four methods evaluated in the ScreenSpot task. The Only image method achieves a success rate of  6.9\%. OmniParser V2 improves significantly to 31.9\%. Screen2AX detected textual representation without hierarchy has a 34.3\% success rate, while the fully formed Screen2AX representation with hierarchy achieves the highest accuracy at 36.6\%, demonstrating the effectiveness of incorporating hierarchy information.}

  \begin{tabular}{lc}
    \toprule
    Method & Success Rate, \% \\
    \midrule
    Only image & 6.9 \\
    OmniParser V2 & 31.9 \\
    Screen2AX without hierarchy & 34.3 \\
    Screen2AX & \textbf{36.6} \\
    \bottomrule
  \end{tabular}
\end{table}

The hierarchical representation generated with Screen2AX provides powerful grounding information for addressing coordinate prediction challenges in macOS interfaces.

\subsubsection{Screen2AX-Task evaluation}

To assess the impact of hierarchical accessibility metadata on the navigation capabilities of autonomous agents, we use the Screen2AX-Task dataset, as described in Section~\ref{sec:datasets}. This benchmark consists of UI navigation tasks designed to evaluate how effectively an agent can interpret various UI representations to perform a given command.

We evaluate four different UI representation formats without image provided: (1) the built-in macOS accessibility metadata, (2) the output of OmniParser V2, (3) a flat list of UI elements detected by Screen2AX with associated descriptions but no hierarchical structure, and (4) the full Screen2AX output, including hierarchical organization. Each format is paired with a consistent prompt template for the agent (see Appendix~\ref{appendix:agent-prompt}).
The agent, powered by GPT-4, is tasked with identifying which UI element should be clicked to perform a specific instruction, phrased as: ``What is the ID of the element that must be clicked to perform the command: \textit{command}?"

Performance is measured by Success Rate, calculated as the proportion of tasks where the agent selects an element whose center lies within the ground-truth bounding box. Table~\ref{tab:tiny_tasks} summarizes the results.

\begin{table}[H]
  \centering
  \caption{Tasks performance evaluation.}
  \Description{A table comparing the success rate of four methods in a task performance evaluation. The methods include Built-in accessibility, OmniParser V2, Screen2AX without hierarchy, and Screen2AX. Built-in accessibility achieves a success rate of 16.9\%, while OmniParser V2 shows improved performance with a success rate of 28.0\%. Screen2AX without hierarchy achieves success rate of 30.7\%, while Screen2AX with hierarchical representation achieves the highest success rate of 33.7\%.}
  \label{tab:tiny_tasks}
  \begin{tabular}{lc}
    \toprule
    Method & Success Rate, \% \\
    \midrule
    Built-in accessibility & 16.9 \\
    OmniParser V2 & 28.0 \\
    Screen2AX without hierarchy & 30.7 \\
    Screen2AX with hierarchy & \textbf{33.7} \\
    \bottomrule
  \end{tabular}
\end{table}

These results demonstrate that our hierarchical representation enhances the agent's performance in screen navigation. Screen2AX with hierarchy achieves a success rate of 33.7\%, which is an absolute improvement of 16.8\% over built-in macOS accessibility and an improvement of 5.7\% compared to OmniPars r v2. Furthermore, incorporating hierarchical structure into Screen2AX raises the success rate by 3\% compared to its non-hierarchical version, underscoring the benefits of structured grouping.

These absolute performance gains indicate that a more structured and detailed representation of UI elements improves the ability of language models like GPT-4 to interpret complex desktop interfaces and execute UI-related tasks effectively.

\section{Qualitative analysis}

In this section, we qualitatively evaluate each core component of our stem. We highlight both its strengths and limitations, particularly in real-world scenarios where UI diversity and ambiguity challenge the robustness of automatic accessibility generation.

\subsection{Accessibility Generation}
\subsubsection{UI Elements Detection}
\label{sec:qual-ui-elements-detection}

\begin{figure*}[ht]
    \centering
    \includegraphics[width=\linewidth]{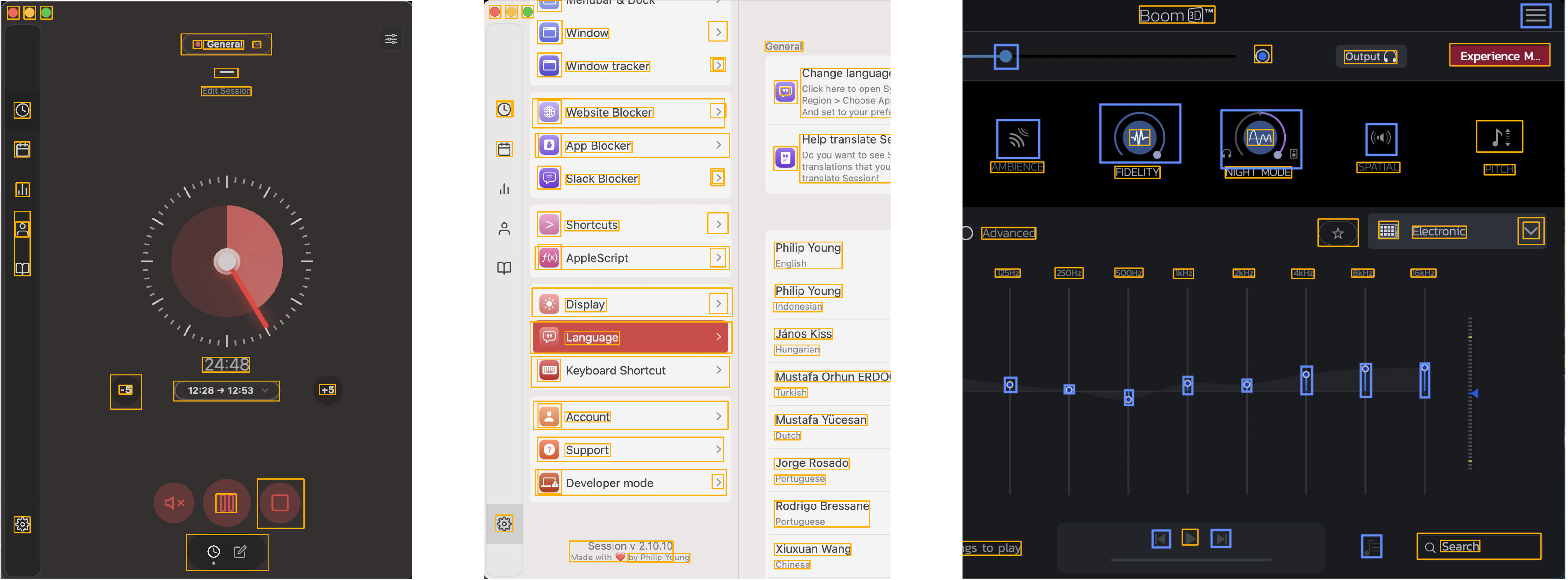}
    \caption{Examples illustrating challenges in achieving perfect prediction accuracy. (Left) An application with rare red buttons at the bottom and an unusual settings button at the top-right, all missed by the model. (Middle) An interface showing inconsistent detection of settings buttons when switching to a light theme. (Right) An atypical macOS application where the model correctly identifies some UI elements (orange outlines) but misses several buttons (blue outlines).}
    \Description{A figure showing three UI screenshots side by side, illustrating failure cases of the model. The left screenshot contains rare red buttons at the bottom and a small settings button at the top right, which the model fails to detect. The middle screenshot shows inconsistent predictions for settings buttons — some are outlined while others are not. The right screenshot features a visually uncommon macOS application where some elements are correctly predicted with orange outlines, but many expected buttons outlined in blue are missed entirely.}
    \label{fig:cornen-cases}
\end{figure*}

Our approach demonstrates strong generalization across common macOS UI components. However, during qualitative analysis, we identified several limitations.

First, the detection model struggles to recognize rare or infrequently occurring UI elements. Because these elements are underrepresented in the training set, the model often fails to generalize their structure and semantics effectively. Figure~\ref{fig:cornen-cases} (left) illustrates a case where rare buttons go undetected.

Second, challenges arise from the quality of developer-provided accessibility data used to construct the Screen2AX-Element dataset. Derived from macOS accessibility hierarchies, this data is often inconsistently labeled and positioned across different applications. For example, semantically similar buttons may be labeled with different or missing roles and depicted in varying ways—one application might denote a button solely by bordering text, while another might include the entire area encompassing the button, icon, and padding. These inconsistencies introduce noise during training, as shown in Figure~\ref{fig:cornen-cases} (middle).

Finally, certain applications feature highly customized or non-standard UI components. In apps like Photoshop or Boom3D, which differ markedly from standard macOS UIs, our model tends to underperform due to visual and semantic divergence from typical design patterns. Figure~\ref{fig:cornen-cases} (right) highlights examples where detection fails on such atypical interfaces.

\begin{figure}[H]
    \centering
    \includegraphics[width=0.12\linewidth]{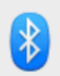}
    \caption{OCR error: Bluetooth icon misinterpreted as the text symbol \textbf{*} by ocrmac.}
    \Description{Standard Bluetooth icon that is misinterpreted as the text symbol \textbf{*} by ocrmac.}
    \label{fig:ocr}
\end{figure}

\begin{table*}[ht]
    \centering
    \caption{Captions from BLIP, OmniParser, and our Screen2AX-tuned BLIP for UI icons. The tuned model generates more precise, context-aware captions aligned with interface semantics.}
    \Description{Table 9 presents a comparison between captions generated by the original BLIP model, OmniParser V2 Florence, and the Screen2AX-tuned BLIP model for a sample of UI  cons. Each row shows a UI icon with a caption corresponding to each model. The original BLIP model gives general visual descriptions, such as a 'Trash can icon' or 'App logo.' OmniParser V2 Florence produces more symbolic or generic functional descriptions like 'T symbol' or 'User profile icon.' In contrast, Screen2AX BLIP provides domain-specific and functionally accurate captions, such as 'Option to delete a file or item' and 'Option to show or reveal the password.' This table highlights the improved semantic alignment of the Screen2AX model with UI functionality.}
    \label{tab:blip-screen2ax-comparison}
    \begin{tabular}{|c|p{5cm}|p{5.5cm}|p{4.2cm}|}
        \hline
        \textbf{Icon} & \textbf{BLIP Caption} & \textbf{OmniParser V2 Florence Caption} & \textbf{Screen2AX BLIP Caption} \\
        \hline
        \raisebox{-0.4\height}{\begin{minipage}[c]{1cm}\centering\vfill\includegraphics[width=1cm]{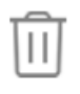}\vfill\end{minipage}} & Trash can icon, typically indicating a trash bin or file deletion option & "T" symbol, typically associated with a function or action & Option to delete a file or item \\
        \hline
        \raisebox{-0.35\height}{\begin{minipage}[c]{1cm}\centering\vfill\includegraphics[width=1cm]{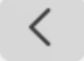}\vfill\end{minipage}} & A white circle with a black arrow, often representing navigation & Pinning function or push pin icon used for fixing items in place & Go to the previous page or step \\
        \hline
        \raisebox{-0.5\height}{\begin{minipage}[c]{1cm}\centering\vfill\includegraphics[width=1cm]{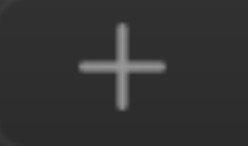}\vfill\end{minipage}} & Black button with a white cross, commonly used to close or dismiss & White plus sign, representing an action to add or create & Option to add a new item or entity \\
        \hline
        \raisebox{-0.35\height}{\begin{minipage}[c]{1cm}\centering\vfill\includegraphics[width=1cm]{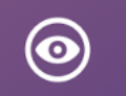}\vfill\end{minipage}} & App logo, often used to represent the app itself & User profile icon or settings for account management & Option to show or reveal the password \\
        \hline
        \raisebox{-0.6\height}{\begin{minipage}[c]{1cm}\centering\vfill\includegraphics[width=1cm]{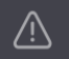}\vfill\end{minipage}} & White alert sign on a black background, indicating a warning or notification & Alert function or notification icon, used for warning & Display an alert or notification message \\
        \hline
    \end{tabular}
\end{table*}
\subsubsection{OCR}

When generating icon captions, we initially use the ocrmac tool \cite{ocrmac} to extract text from bu tons. Although ocrmac typically performs well, it can struggle when an icon closely resembles a text symbol. For example, as shown in Figure~\ref{fig:ocr}, the tool mistakenly detects a star symbol instead of the intended Bluetooth symbol.

\subsubsection{Icon captioning}
\label{sec:icon-captioning-analysis}

We compare our fine-tuned BLIP model against the original BLIP, and the fine-tuned Florence model from OmniParser V2 \cite{lu2024omniparserpurevisionbased}, as shown in Table~\ref{tab:blip-screen2ax-comparison}. While the original BLIP model is capable of recognizing visual patterns, it lacks an understanding of UI-specific semantics and terminology. Additionally, the icon captioning model from OmniParser V2 struggles with common UI elements and often produces inaccurate or overly generic descriptions. In contrast, our fine-tuned BLIP model demonstrates a significantly improved understanding of UI concepts, producing more precise and contextually appropriate captions for interface components. These results highlight the advantage of domain-specific fine-tuning for UI understanding tasks.

\begin{table}[ht!]
\centering
\caption{Analysis of BLIP failure cases. One likely cause is the limited visual complexity of mobile UI icons used during fine-tuning, which may not generalize well to more intricate desktop UI elements.}
\Description{Table 10 shows examples of failure cases from the Screen2AX BLIP model. Each row includes an icon, the model’s generated caption, and the expected ground-truth caption. The model incorrectly identifies a 'share' icon as 'go to home page,' a 'data storage' icon as 'go to the previous,' and a 'create table' icon as 'go to app.' The table analysis suggests that these failures are likely due to the limited visual complexity of mobile UI icons used during fine-tuning, which may not generalize well to more complex desktop UI elements.}
\label{tab:blip-failures}
\begin{tabular}{|c|p{3.5cm}|p{3cm}|}
    \hline
    \textbf{Icon} & \textbf{Screen2AX BLIP Caption} & \textbf{Expected Caption (Ground-Truth)} \\
    \hline
    \raisebox{-0.5\height}{\begin{minipage}[c]{1.2cm}\centering\vfill\includegraphics[width=1cm]{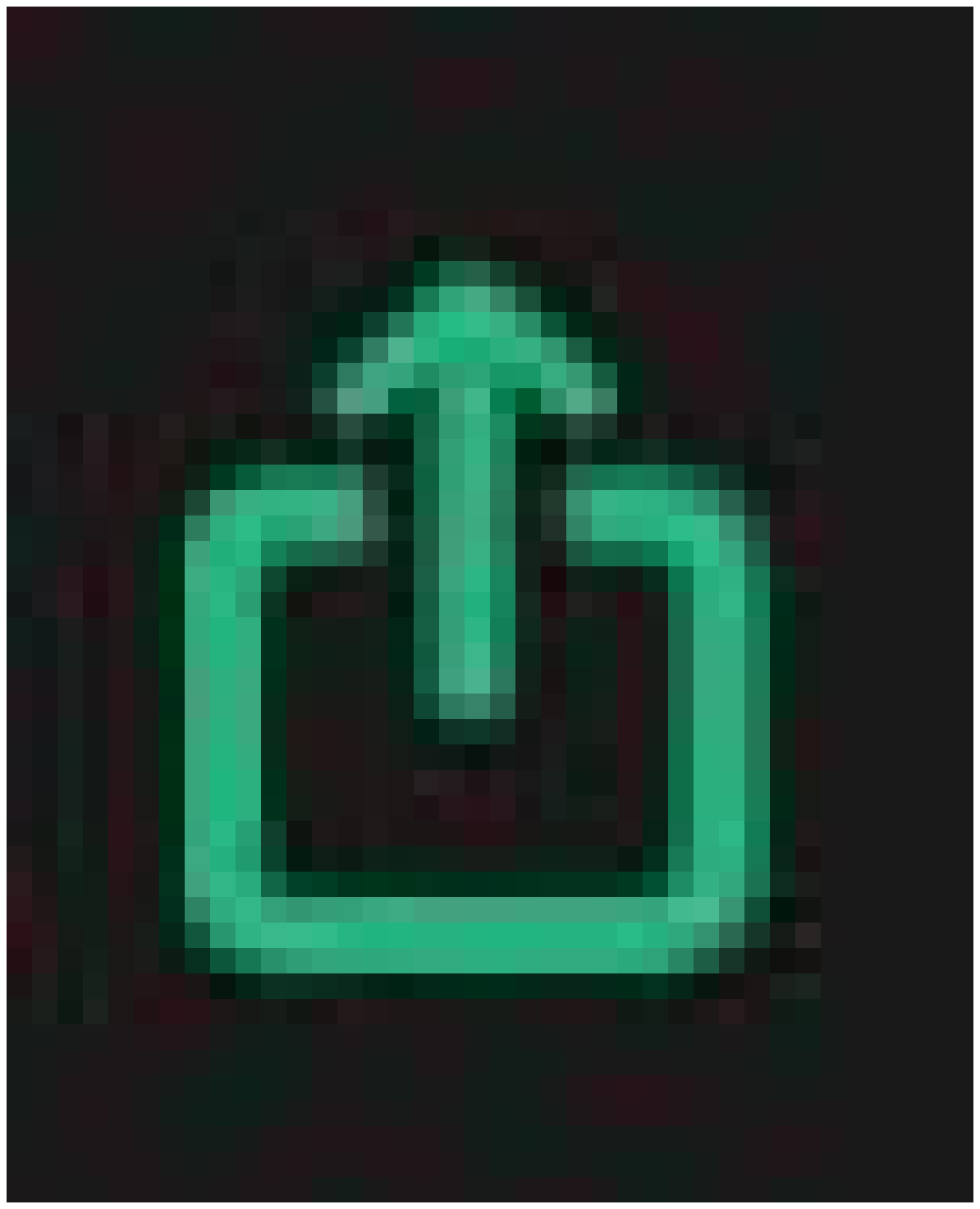}\vfill\end{minipage}} & go to home page & share \\
    \hline
    \raisebox{-0.25\height}{\begin{minipage}[c]{1.2cm}\centering\vfill\includegraphics[width=1cm]{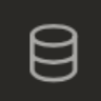}\vfill\end{minipage}} & go to the previous & data storage \\
    \hline
    \raisebox{-0.4\height}{\begin{minipage}[c]{1.2cm}\centering\vfill\includegraphics[width=1cm]{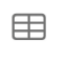}\vfill\end{minipage}} & go to app & create table \\
    \hline
\end{tabular}
\end{table}

Next, we examine the failure cases of our fine-tuned BLIP model, some of which involve uncommon or desktop-specific icons. Examples of these cases are shown in Table~\ref{tab:blip-failures}.

These failures are likely due to the model being fine-tuned on a mobile UI dataset, which differs significantly from macOS interfaces. In particular, icons such as “Data Storage” or “Create Table” are either rare or nonexistent in mobile environments, making it difficult for the model to generalize effectively to these desktop-specific elements.
    
\subsubsection{UI Groups Detection}
Similar to UI Elements Detection (Section~\ref{sec:qual-ui-elements-detection}), the primary challenge in UI group detection stems from inconsistent labeling practices across applications, which is reflected in our dataset. Since grouping strategies vary significantly between developers, the resulting dataset exhibits high variability in how elements are structured and clustered into groups.

Additionally, as introduced in Section~\ref{sec:datasets}, the distribution of the number of groups per screenshot is not uniform. This inconsistency introduces challenges for learning a generalizable grouping strategy.
Due to this uneven distribution and lack of standardized grouping conventions, our model tends to predict different grouping techniques across different interfaces — and sometimes even within the same interface.

\subsection{Qualitative analysis of Screen2AX-Task}

We analyzed how an AI agent performs tasks using the accessibility structure generated by Screen2AX. 
Since the agent operates solely based on the JSON accessibility representation, its performance depends critically on the accuracy of both the UI element detection (bounding boxes) and the element descriptions generated by our BLIP model.

\begin{figure}
    \centering
    \includegraphics[width=\linewidth]{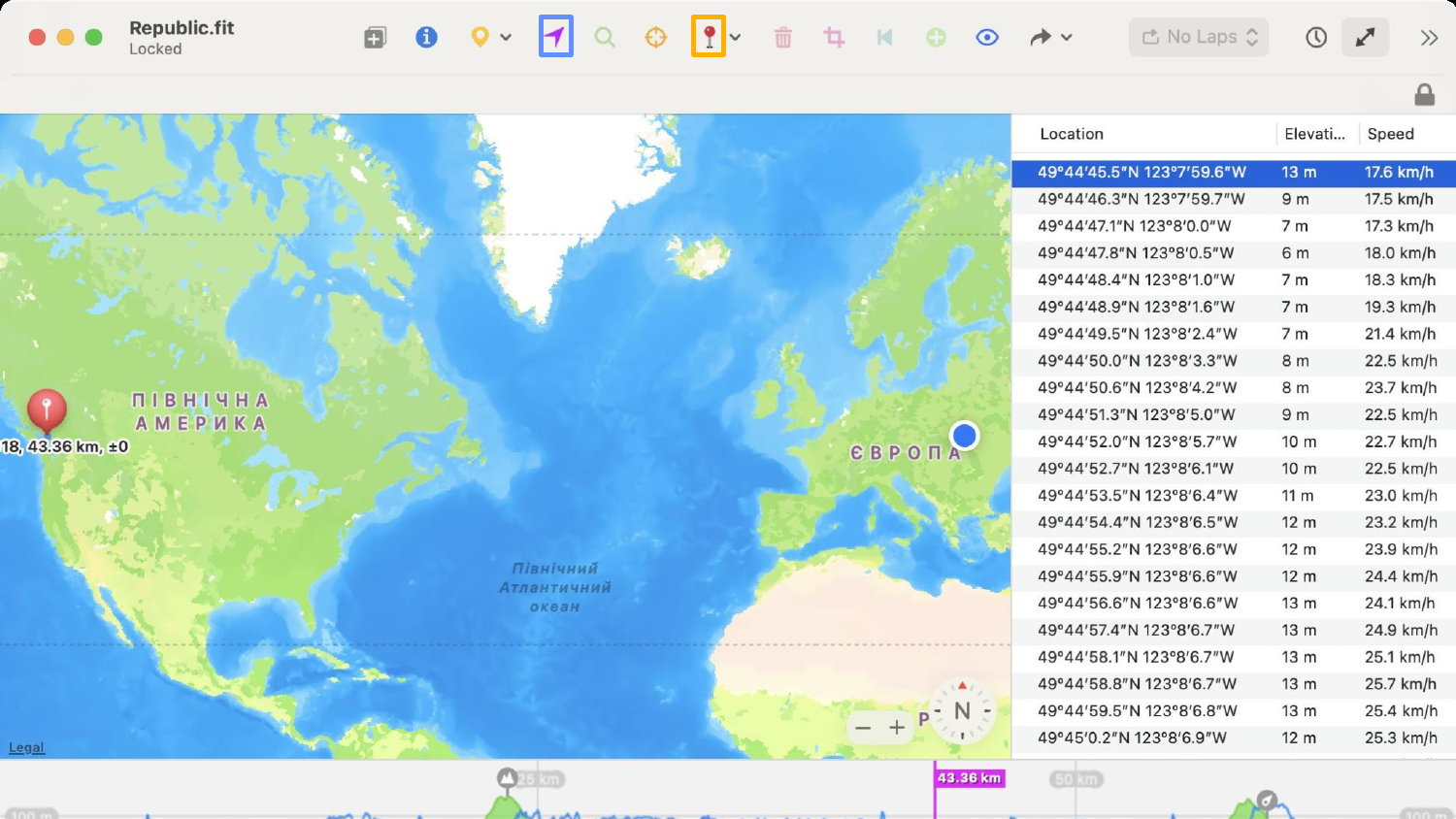}
    \caption{Example of failure of task completion by AI agent. The task is "add location mark r". The orange box indicates the ground truth button that needs to be clicked, while the blue one shows the mispredicted by the agent. The failure is due to our BLIP caption for ground truth being "go to next".}
    \Description{Figure 10 illustrates a failure case in task completion by an AI agent. The task is to add a location marker on a map interface. The image shows a map with UI controls, where an orange box highlights the correct button to add the marker, and a blue box highlights the incorrect button clicked by the agent. The failure is attributed to a mismatch in the BLIP-generated caption, where the caption for the correct button was mistakenly interpreted as 'go to next' instead of 'add location marker.'}
    \label{fig:analysis-tasks}
\end{figure}

Figure~\ref{fig:analysis-tasks} illustrates a typical failure case where task execution was unsuccessful. This failure primarily stems from BLIP's inability to accurately caption elements within complex interfaces, a limitation previously discussed in Section~\ref{sec:icon-captioning-analysis}. When faced with densely populated UIs, the captioning model struggles to provide sufficiently descriptive and accurate element labels, making it difficult for the agent to identify the correct interactive elements for completing the assigned task.

\section{Discussion \& Limitations}

Our results demonstrate that Screen2AX significantly advances autonomous agent performance in UI navigation and task execution, outperforming existing state-of-the-art screen parsing methods. By generating structured, hierarchical accessibility metadata for complex desktop interfaces, our approach fills a critical gap in desktop accessibility research.

When comparing our results with prior work, Screen2AX shows particular strength in the hierarchical representation of UI elements, which contributes to more inclusive computing experiences for individuals who rely on assistive technologies. 
The semantic grouping of elements represents an advance over non-hierarchical structural approaches seen in previous research. 

Despite our approach's strong performance, several limitations remain. First, the quality of our dataset is constrained by its dependence on developer-provided accessibility annotations, which can be incomplete, inconsistent, or error-prone. While our dataset is among the largest available for macOS interfaces, it may not fully capture the diversity of macOS application designs and UI patterns.

Another challenge arises from the domain gap in UI element captioning. Our captioning model was initially fine-tuned on icon sets derived from mobile UIs, which introduces a potential domain gap, as many macOS icons are more complex or visually distinct from their mobile counterparts. As a result, the model may struggle with uncommon or desktop-specific icons (e.g., “Data Storage” or “Create Table”), limiting generalization. Expanding the dataset to include a wider variety of desktop-specific icons and associated captions would likely improve robustness and accuracy.

Another limitation of our approach lies in its inference speed, which may not meet the latency requirements of real-time applications. Future work could explore performance optimizations such as model pruning, quantization, or the adoption of more efficient architectures to improve inference time without sacrificing accuracy.

Lastly, a promising direction for future research lies in classifying UI element groups by semantic roles—such as toolbars, navigation bars, or side panels. This semantic structuring could further improve the agent's navigation performance and benefit users relying on assistive technologies.

\section{Conclusions}

In this paper, we introduced Screen2AX, a novel vision-based pipeline for generating rich, hierarchical accessibility metadata from a single macOS application screenshot. Our system uses computer vision and deep learning techniques to localize, classify, describe, and semantically group UI elements directly from screenshots, automating the traditionally labor-intensive annotation process.

The evaluation results demonstrate that our approach improves the accuracy of UI element detection and description and produces a more structured and comprehensive representation of the user interface, thereby enhancing both human-centric assistive technologies for desktop usage and AI-driven autonomous agents.

The findings underscore the potential of vision-based methods to overcome common accessibility challenges, such as misclassified elements, missing metadata, and inconsistent UI structures. 

Looking forward, this research opens several promising directions. Integrating Screen2AX in development could enable real-time accessibility validation during the application design process. Additionally, extending our approach to other operating systems and application types represents an important next step toward increasing the availability of accessibility metadata.
Ultimately, this approach contributes to more inclusive user experiences, ensuring that assistive technologies can better interpret and interact with dynamic user interfaces.

\bibliographystyle{ACM-Reference-Format}
\bibliography{sample}

\include{appendix}

\end{document}

%% file: appendix.tex
\appendix
\section{Accessibility}
\label{appendix:accessibility}

\subsection{Accessibility Elements Classification}

The elements in the built-in macOS accessibility metadata are divided into 52 classes: AXGroup, AXOpaqueProviderGroup, AXRadioGroup, AXSplitGroup, AXTabGroup, AXToolbar, AXWebArea, AXOutline, AXSplitter, AXSheet, AXBrowser, AXPopover, AXGrid, AXGrowArea, AXList, AXTable, AXScrollArea, AXWindow, AXPage, AXStaticText, AXHeading, AXLink, AXCheckBox, AXRadioButton, AXSlider, AXComboBox, AXScrollBar, AXButton, AXPopUpButton, AXMenuButton, AXDisclosureTriangle, AXIncrementor, AXColorWell, AXIncrementorArrow, AXTextField, AXTextArea, AXDateTimeArea, AXCell, AXImage, AXBusyIndicator, AXUnknown, AXGenericElement, AXRuler, SWTComposite, JavaAxIgnore, AXMenuItem, AXMenu, AXMenuBar, AXListMarker, AXValueIndicator, AXProgressIndicator.

\begin{figure}[H]
    \centering
    \includegraphics[width=\linewidth]{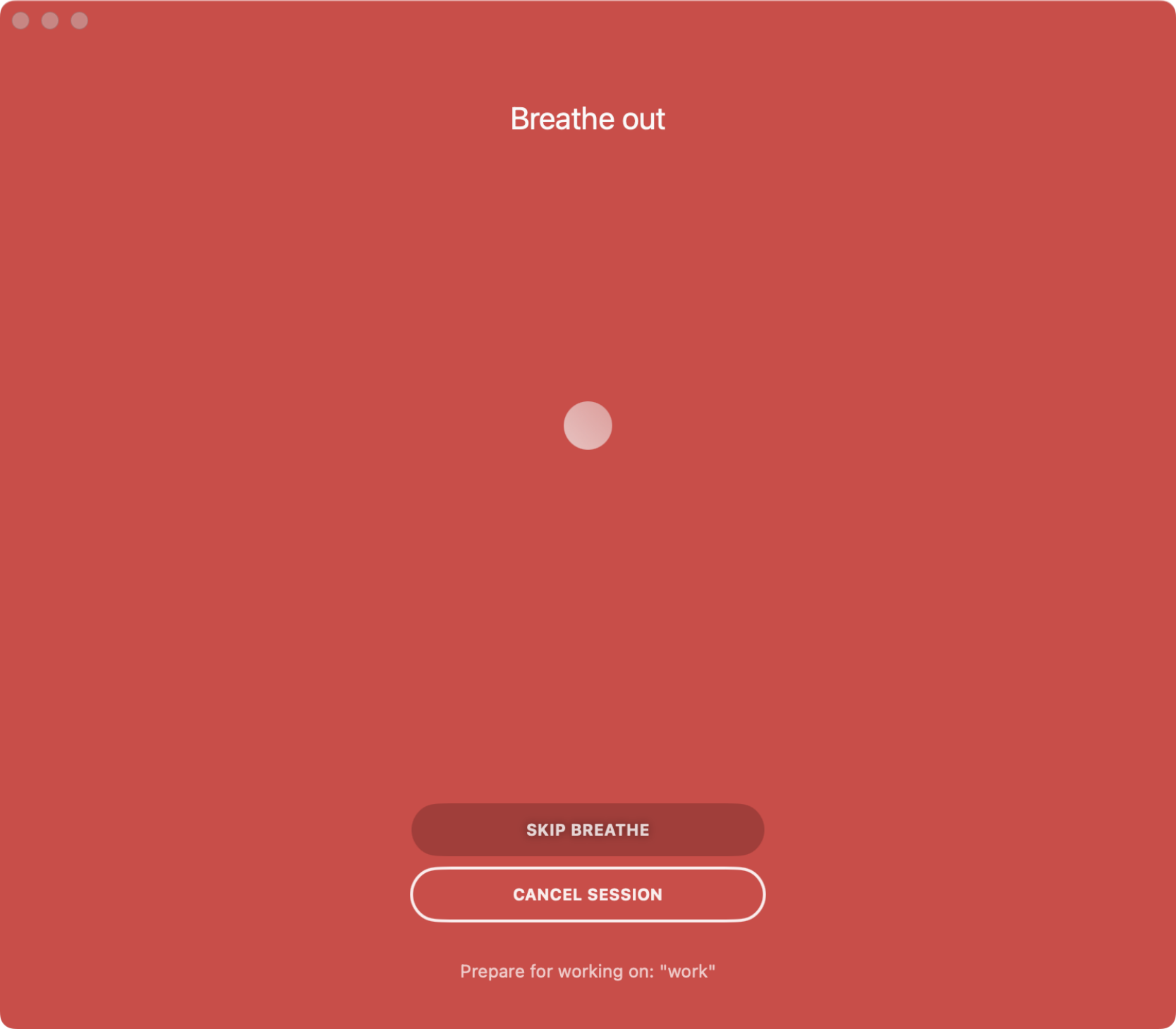}
    \caption{Screenshot of the UI from macOS application "Session Pomodoro Focus Timer".}
    \Description{Screenshot of the Session Pomodoro Focus Timer app. The corresponding accessibility tree is shown below.}
    \label{fig:tree-example}
\end{figure}

\subsection{Accessibility Tree Example} \label{appendix:accessibility-tree}
Figure~\ref{fig:tree-example} shows the screenshot of the UI, and Listing~\ref{lst:accessibility-tree} presents the corresponding accessibility hierarchy tree in JSON format.

\lstdefinelanguage{json}{
  morestring=[b]",
  morecomment=[l]{//},
  stringstyle=\color{blue},
  commentstyle=\color{gray},
  morekeywords={true,false,null},
  keywordstyle=\color{red},
}

\lstdefinestyle{jsonstyle}{
  language=json,
  basicstyle=\ttfamily\scriptsize,
  numbers=left,
  numberstyle=\tiny\color{gray},
  stepnumber=1,
  numbersep=6pt,
  backgroundcolor=\color{white},
  showstringspaces=false,
  breaklines=true,
  breakindent=1em,
  frame=single,
  rulecolor=\color{black},
  tabsize=2,
  captionpos=b,
  xleftmargin=1.5em
}

\begin{lstlisting}[style=jsonstyle, caption={Excerpt of accessibility tree for the Session Pomodoro Focus Timer}, label={lst:accessibility-tree}]
{
  "name": "Session Pomodoro Focus Timer",
  "role": "AXWindow",
  "description": null,
  "role_description": "standard window",
  "value": null,
  "children": [
    {
      "name": null,
      "role": "AXGroup",
      "description": null,
      "role_description": "group",
      "value": null,
      "children": [
        { "name": null, "role": "AXButton", "description": "Close", "role_description": "button", "value": null, "children": [], "bbox": [844, -296, 40, 40], "visible_bbox": null },
        { "name": null, "role": "AXStaticText", "description": null, "role_description": "text", "value": "Breathe out", "children": [], "bbox": [694, 104, 212, 50], "visible_bbox": [694, 104, 212, 50] },
        { "name": null, "role": "AXButton", "description": "clock", "role_description": "button", "value": null, "children": [], "bbox": [0, 200, 94, 96], "visible_bbox": [0, 200, 94, 96] },
        { "name": null, "role": "AXButton", "description": "bar-chart-2", "role_description": "button", "value": null, "children": [], "bbox": [0, 296, 94, 96], "visible_bbox": [0, 296, 94, 96] },
        { "name": null, "role": "AXButton", "description": "user", "role_description": "button", "value": null, "children": [], "bbox": [0, 392, 94, 96], "visible_bbox": [0, 392, 94, 96] },
        { "name": null, "role": "AXButton", "description": "book-open", "role_description": "button", "value": null, "children": [], "bbox": [0, 488, 94, 96], "visible_bbox": [0, 488, 94, 96] },
        { "name": null, "role": "AXButton", "description": "Skip breathe", "role_description": "button", "value": null, "children": [], "bbox": [560, 1092, 480, 72], "visible_bbox": [560, 1092, 480, 72] },
        { "name": null, "role": "AXButton", "description": "Cancel Session", "role_description": "button", "value": null, "children": [], "bbox": [560, 1180, 480, 72], "visible_bbox": [560, 1180, 480, 72] },
        { "name": null, "role": "AXButton", "description": "settings", "role_description": "button", "value": null, "children": [], "bbox": [0, 1224, 94, 96], "visible_bbox": [0, 1224, 94, 96] },
        { "name": null, "role": "AXStaticText", "description": null, "role_description": "text", "value": "Prepare for working on: \"work\"", "children": [], "bbox": [624, 1306, 348, 30], "visible_bbox": [624, 1306, 348, 30] }
      ],
      "bbox": [0, 0, 1600, 1400],
      "visible_bbox": [0, 0, 1600, 1400]
    },
    { "name": null, "role": "AXButton", "description": null, "role_description": "close button", "value": null, "children": [], "bbox": [14, 12, 28, 32], "visible_bbox": [14, 12, 28, 32] },
    { "name": null, "role": "AXButton", "description": null, "role_description": "full screen button", "value": null, "children": [], "bbox": [94, 12, 28, 32], "visible_bbox": [94, 12, 28, 32] },
    { "name": null, "role": "AXButton", "description": null, "role_description": "minimize button", "value": null, "children": [], "bbox": [54, 12, 28, 32], "visible_bbox": [54, 12, 28, 32] }
  ],
  "bbox": [0, 0, 1600, 1400],
  "visible_bbox": [0, 0, 1600, 1400]
}
\end{lstlisting}

\section{GPT-4 Classification Prompt}
\label{appendix:classification-prompt}

Listing~\ref{lst:gpt-classification} shows the prompt used to classify the detected UI elements using GPT-4.

\begin{lstlisting}[basicstyle=\small\ttfamily, frame=single, breaklines=true, label={lst:gpt-classification}]
In this task, your goal is to analyze a UI screenshot of the {app_name} application on macOS and given segments on this screenshot, delineating each with a type, description and text. The screenshot will feature various segments highlighted in green boxes, labeled from 1 to {boxes_number}, sorted by y-coordinate. The labels are always white numbers with a black outline.
Your task is to identify the type of each element from the following options: AXButton, AXStaticText, AXTextArea, AXGroup, AXImage, AXLink, WRONGLY_SEGMENTED, UNCLEAR.
<important>It is crucial to use the "WRONGLY_SEGMENTED" type if the segmentation of an element is incorrect and the element is not a UI element. Additionally, use the "UNCLEAR" type if the type of element cannot be determined with 100% certainty.</important>
For each identified element, provide a description that correlates with its functionality. If the element type is AXImage, write an alternative text describing the image content. Include text from the box if it is presented.
Ensure that your responses are structured in a JSON format, allowing for easy identification and mapping of elements to their respective types and descriptions.

Example:
```json
[
    {
        "id": 1,
        "type": "AXButton",
        "description": "A clickable button that pauses the currently playing podcast episode.",
        "text": ""
    },
    {
        "id": 2,
        "type": "AXStaticText",
        "description": "A non-editable text field displaying the title of the currently playing podcast episode.",
        "text": "Back to Black"
    },
    {
        "id": 3,
        "type": "WRONGLY_SEGMENTED",
        "description": "The segmentation is incorrect as this element does not correspond to any UI component.",
        "text": ""
    },
    {
        "id": 4,
        "type": "AXImage",
        "description": "A podcast poster featuring Her Majesty Queen Elizabeth II with her dogs driving a car.",
        "text": "Her Majesty Queen Elizabeth II with her dogs driving the car"
    }
]
```
\end{lstlisting}

\section{GPT-Measured Accuracy Prompt}
\label{appendix:gpt-accuracy-prompt}

To assess the semantic accuracy of predicted UI icon descriptions, we use GPT-4 to perform a pairwise comparison between the ground truth and the predicted captions. The model is prompted to return 1 if the two captions convey the same meaning, and 0 otherwise. This binary evaluation captures whether the essential semantics of the icon have been preserved, even when the wording differs.

\begin{lstlisting}[basicstyle=\small\ttfamily, frame=single, breaklines=true, label={lst:description-similarity-prompt}]
You are given two strings that corresponds to the description of the button. 
Do the following two answers have the same meaning (correspond to the same button that have the same functionality)? 
Answer with 1 (yes) or 0 (no). 
Description 1: {description_1} 
Description 2: {description_2}
\end{lstlisting}

\section{Screen2AX-Task Analysis}
\label{appendix:task-analysis}
To understand the characteristics and challenges of the Screen2AX-Task dataset, we conducted a comprehensive analysis of its composition and structure. This analysis serves three key purposes: (1) to verify that the dataset reflects realistic distributions found in modern user interfaces, (2) to identify potential biases that might influence model performance, and (3) to establish baseline expectations for how different UI elements might be represented in the dataset.

\begin{figure}[H]
    \centering
    \includegraphics[width=\linewidth]{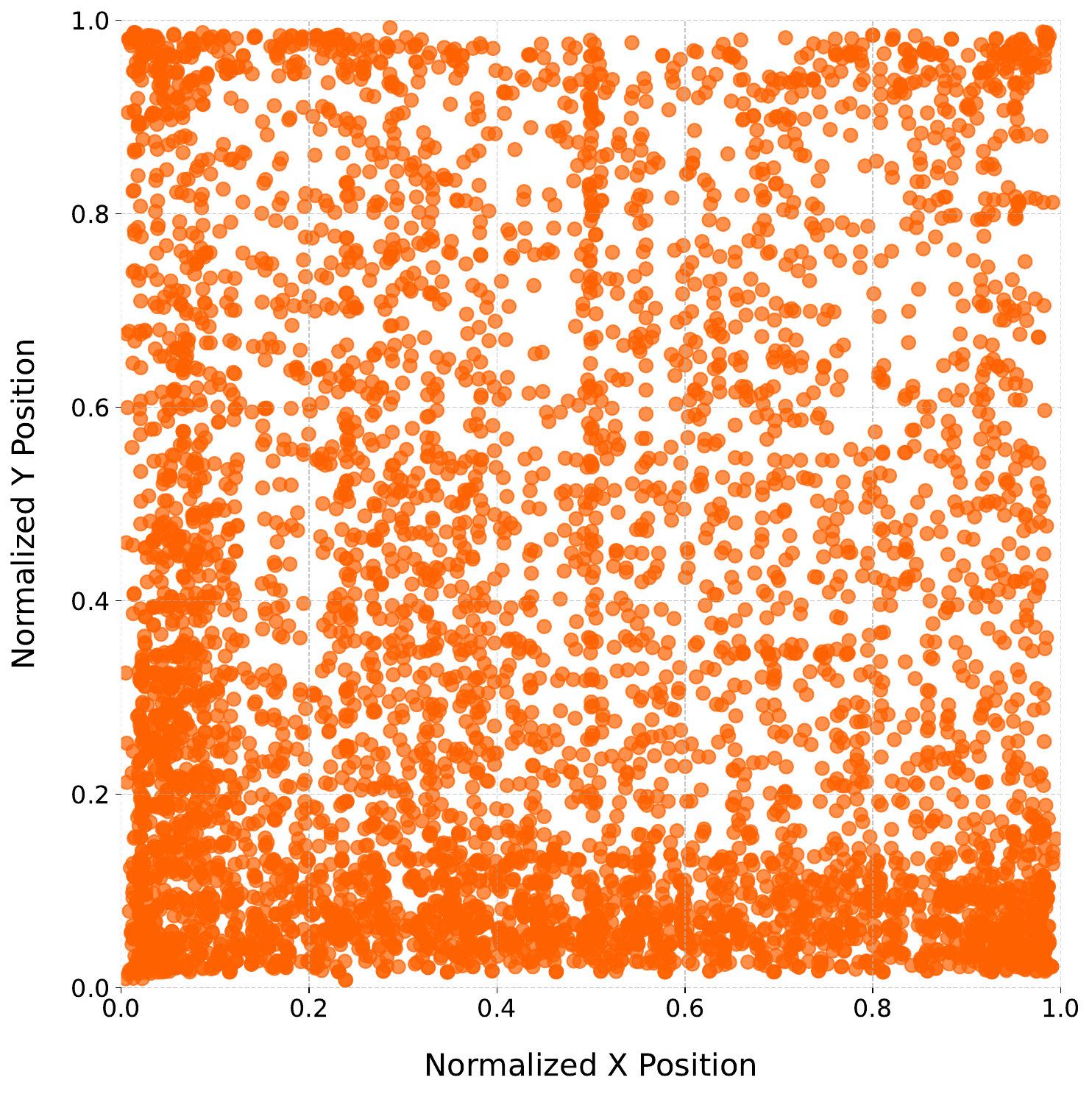}
    \caption{Spatial density of UI elements. The plot shows the normalized positions of all elements in the dataset. Note the concentration of elements along screen edges and bottom areas, reflecting common UI design patterns such as navigation bars, status indicators, and action buttons.}
    \Description{A scatter plot showing the distribution of UI elements across normalized screen coordinates. Each point represents an element position, with x-axis showing horizontal position (0 to 1) and y-axis showing vertical position (0 to 1). Points appear in orange and are distributed across the entire plot, with higher densities visible along the edges and particularly along the bottom of the screen. The distribution is non-uniform but covers the entire coordinate space, with some clustering patterns visible along the borders.}
    \label{fig:element_positions}
\end{figure}

\begin{figure}[H]
    \centering
    \includegraphics[width=\linewidth]{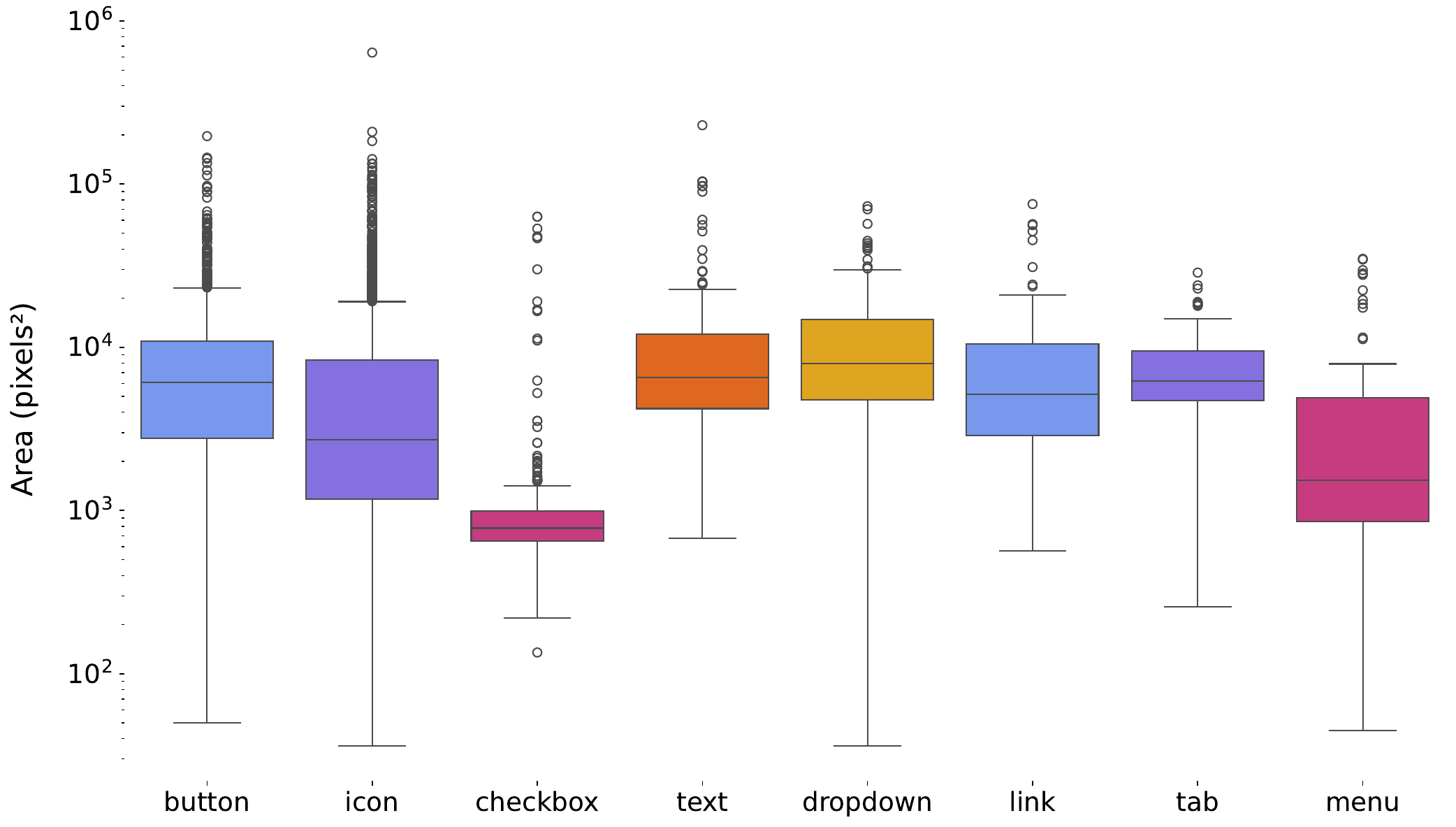}
    \caption{Size distribution by element type. Boxplots show the area (in pixels$^2$) for each element type on a logarithmic scale. Note the substantial size variation within and between element types, with checkboxes and menus being consistently smaller than text elements and dropdowns.}
    \Description{A boxplot showing size distribution across eight UI element types: button, icon, checkbox, text, dropdown, link, tab, and menu. The y-axis uses a logarithmic scale from 10^2 to 10^6 pixels squared. Checkboxes have the smallest median size (around 800 pixels squared), while dropdowns and text elements have the largest (around 8,000-10,000 pixels squared). Icons and buttons show the widest range with many outliers extending to 10^5 pixels squared. The visualization uses different colors for each element type and shows quartiles, medians, and outliers.}
    \label{fig:element_sizes}
\end{figure}

\begin{figure}[H]
    \centering
    \includegraphics[width=\linewidth]{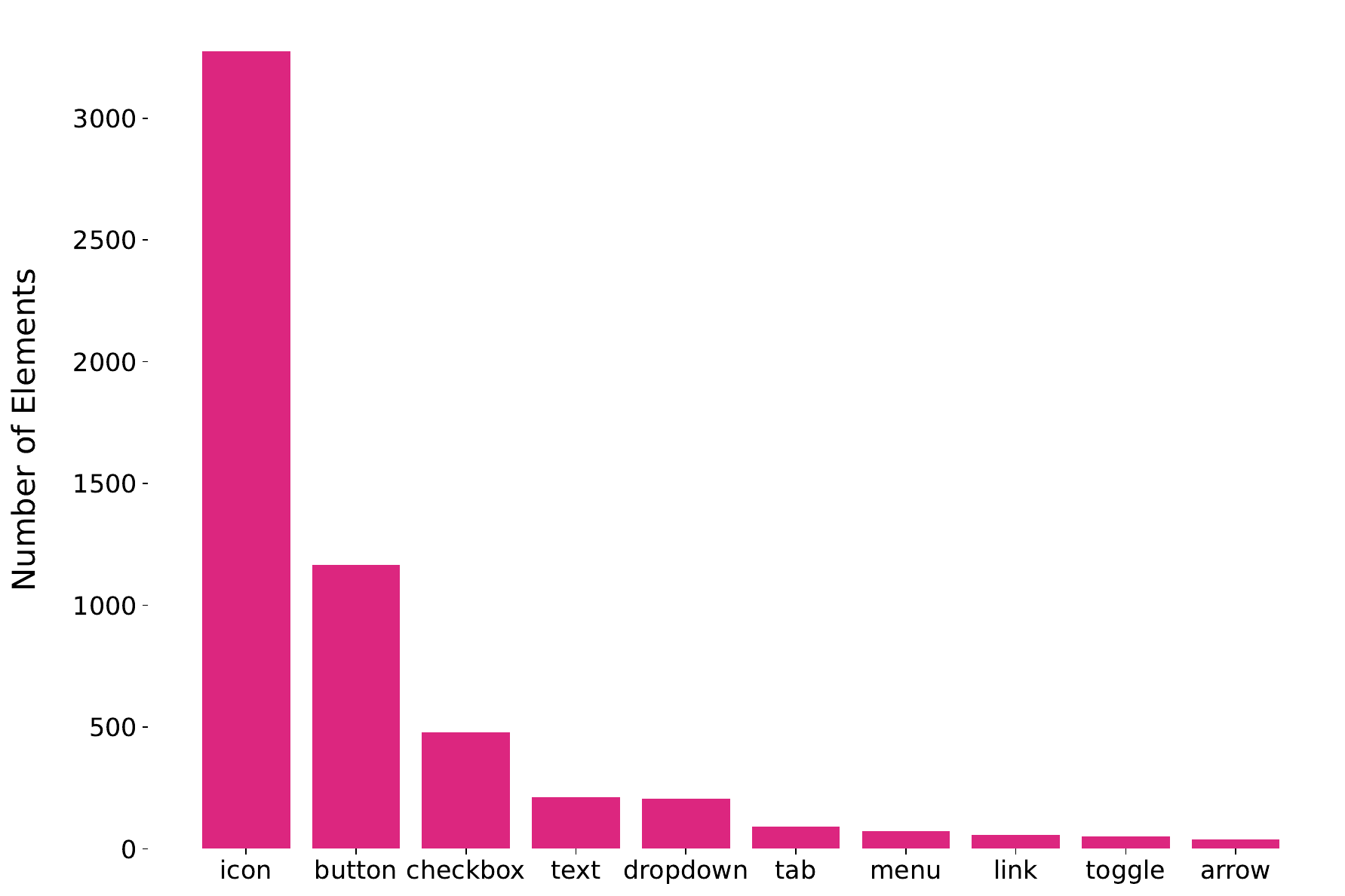}
    \caption{Distribution of element types in the dataset. The plot highlights the significant class imbalance, with icons comprising 55.3\% of all elements, followed by buttons (19.7\%) and checkboxes (8.1\%).}
    \Description{A bar chart showing the frequency distribution of UI element types. The x-axis lists ten element types (icon, button, checkbox, text, dropdown, tab, menu, link, toggle, arrow) while the y-axis shows the count from 0 to 3000. Icons dominate with approximately 3000 instances, followed by buttons with about 1200 instances, and checkboxes with roughly 500 instances. The remaining element types (text, dropdown, tab, menu, link, toggle, and arrow) each have fewer than 250 instances, creating a long-tail distribution. All bars are shown in a magenta color.}
    \label{fig:element_distribution}
\end{figure}

The following visualizations examine three fundamental aspects of the dataset: spatial distribution of elements across interface screens (Figure \ref{fig:element_positions}), size variations across different element types (Figure \ref{fig:element_sizes}), and class distribution among element categories (Figure \ref{fig:element_distribution}). 

Figure~\ref{fig:element_positions} demonstrates that while the distribution of element position is non-uniform—with higher densities along edges and bottom areas reflecting common UI conventions—the dataset successfully captures elements across the entire screen space. This positional diversity ensures the benchmark tests models' ability to identify UI elements regardless of their location, providing a realistic evaluation scenario that mirrors actual interface designs.

Figure~\ref{fig:element_sizes} illustrates the substantial size variation across UI element types. Checkboxes are consistently small (median ~800 pixels$^2$), while dropdowns and text elements are significantly larger (median ~8,000-10,000 pixels$^2$). This heterogeneity, including outliers spanning multiple orders of magnitude for buttons and icons, presents a realistic challenge for UI understanding models. The logarithmic scale highlights that robust models must handle not only different element types but also dramatic size variations within the same category—a common characteristic of real-world interfaces.

Figure~\ref{fig:element_distribution} illustrates the natural class imbalance in UI elements. This distribution mimics real-world interfaces, with icons and buttons dominating while elements like tabs, menus, and toggles form a long tail. Such imbalance presents a meaningful challenge for robust models, which must learn to recognize both common and rare element types with comparable accuracy—a critical capability for systems that aim to interact with diverse UI designs.

\section{Agent Prompt}
\label{appendix:agent-prompt}

The Listing~\ref{lst:agent-prompt} presents the prompt used by GPT-4 to perform UI-related tasks using the accessibility metadata provided.

\begin{lstlisting}[basicstyle=\small\ttfamily, frame=single, breaklines=true, label={lst:agent-prompt}]
You are given a list of UI elements in JSON format, each with a unique numeric ID and accessibility attributes.
Your task is to determine which UI element should be clicked to perform a specific action.
Return only the numeric ID of the element that corresponds to the action.
Do not explain or output anything else.
Accessibility JSON: {accessibility_json}
Action: {action}
Which element should be clicked?
\end{lstlisting}

\section{Confusion matrices}
\label{appendix:confusion-matrices}

\begin{figure}[H]
    \centering
    \includegraphics[width=\linewidth]{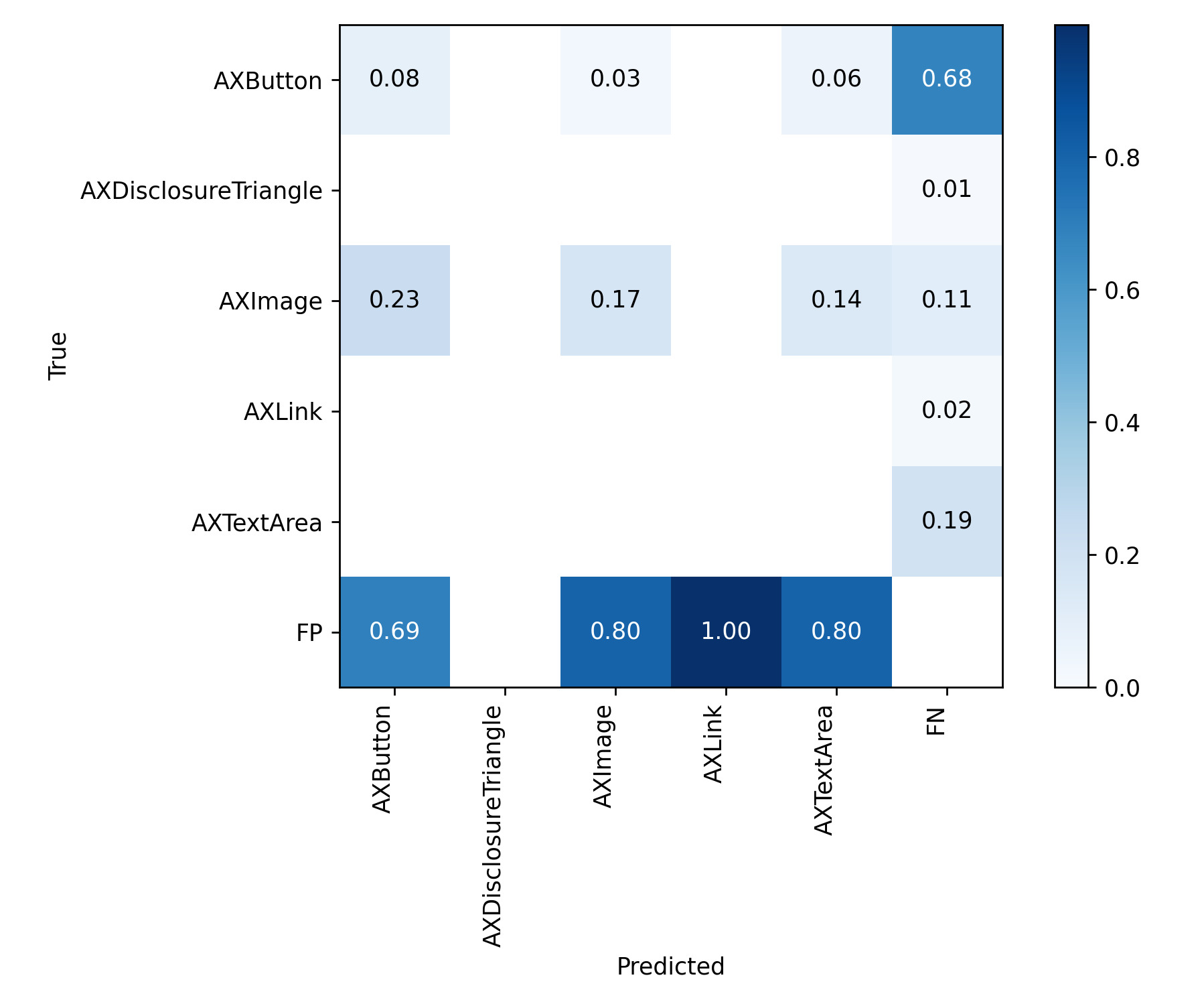}
    \Description{This confusion matrix illustrates the performance of the MSER+YOLOv11 pipeline in detecting and classifying macOS accessibility UI elements. The matrix is row-normalized, where each row corresponds to the true class and each column to the predicted class. The diagonal elements represent correct classifications, while off-diagonal values indicate misclassifications. High false positive (FP) rates are observed across several classes—particularly for AXLink and AXImage—suggesting over-detection. Conversely, false negatives (FN) are notably high for AXButton, indicating under-detection. These results highlight specific weaknesses in class discrimination and the need for refinement in the model or preprocessing stage. Also, some classes like AXDisclosureTriangle and AXLink remained undetected.}
    \caption{Normalized confusion matrix of the MSER+YOLOv11 approach for macOS accessibility element detection.}
    \label{fig:deterministic_cm}
\end{figure}

\begin{figure}[H]
    \centering
    \includegraphics[width=\linewidth]{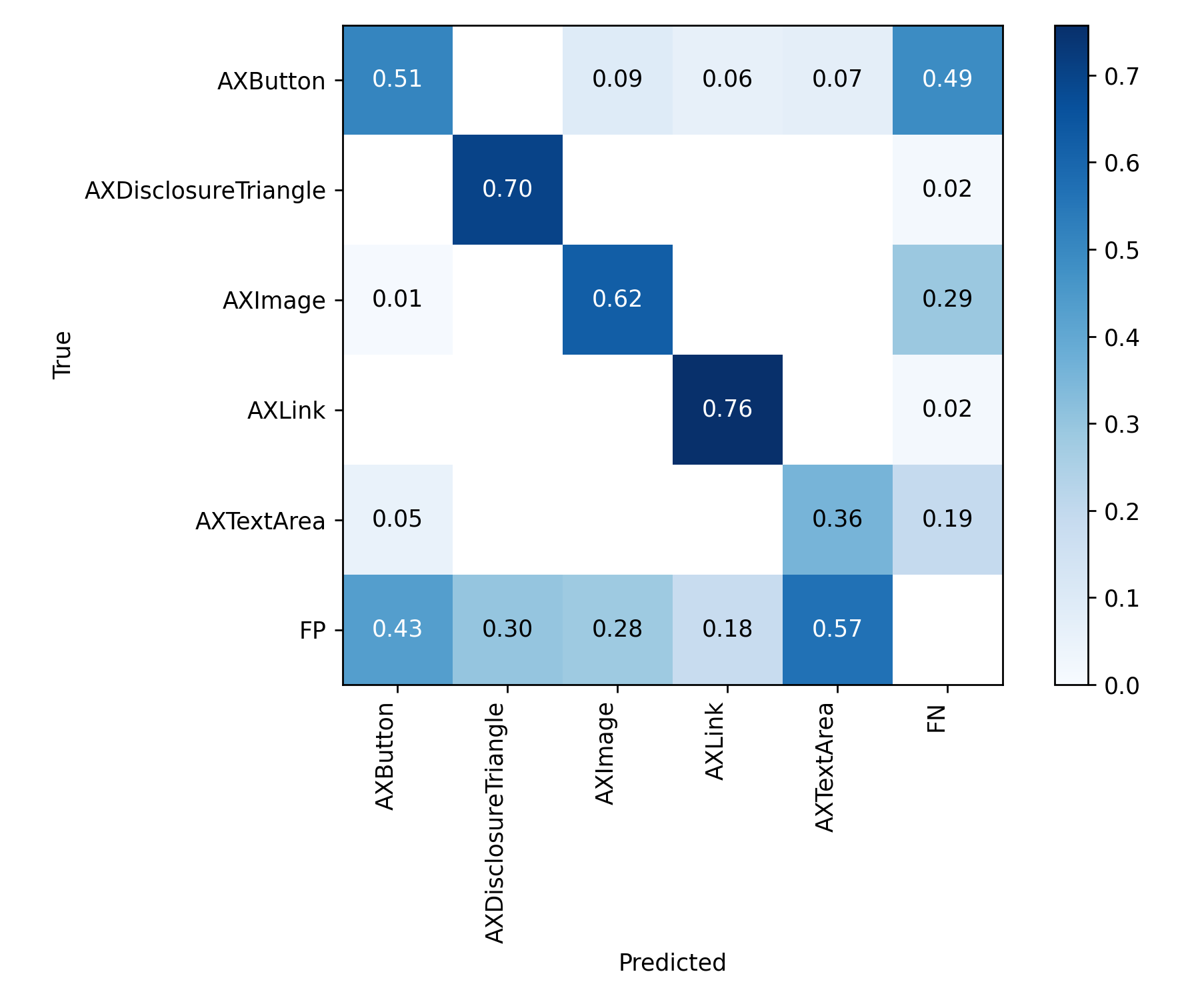}
    \Description{This confusion matrix presents the normalized classification performance of the YOLOv11 model on macOS UI accessibility elements. The diagonal values reflect correct predictions, while off-diagonal entries denote misclassifications. YOLOv11 demonstrates strong class discrimination for AXLink (0.76), AXDisclosureTriangle (0.70), and AXImage (0.62). Notably, AXButton shows a near-equal balance between correct detections (0.51) and false negatives (0.49), indicating inconsistent detection. Although false positives remain present—especially for AXTextArea and AXButton—they are generally lower than in the MSER+YOLOv11 approach, suggesting an overall improvement in precision. This matrix highlights YOLOv11’s effectiveness in isolating specific element types, while still leaving room for improvement in handling ambiguous or overlapping UI features.}
    \caption{Normalized confusion matrix of YOLOv11 for macOS accessibility element detection.}
    \label{fig:yolo_cm}
\end{figure}

The figures show the confusion matrices of UI element detection approaches at the classification stage. Specifically, Figure~\ref{fig:deterministic_cm} presents the confusion matrix for the MSER+YOLOv11 approach, while Figure~\ref{fig:yolo_cm} shows the confusion matrix for YOLOv11 element detection alone. The results suggest that YOLOv11 is significantly more accurate in classifying UI element types.

\begin{figure}[H]
    \centering
    \includegraphics[width=\linewidth]{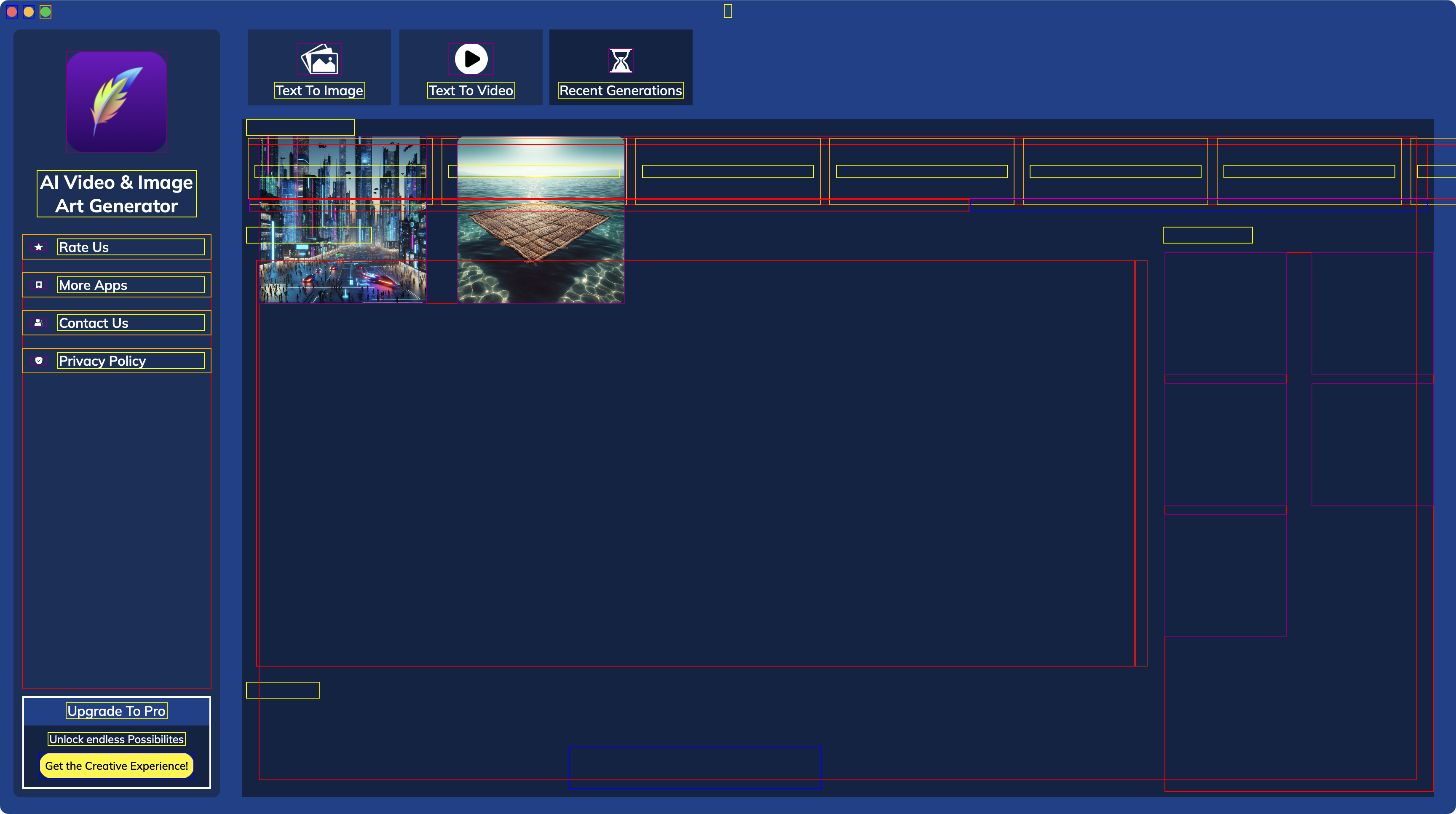}
    \Description{A user interface screenshot showing bounding boxes for elements that are not visually present in the application. These "invisible" elements, although included in the accessibility metadata, do not correspond to any actual UI components, making navigation confusing.}
    \caption{An example screenshot of UI from an application where the built-in accessibility contains invisible elements.}
    \label{fig:invisible-elements}
\end{figure}

\begin{figure}[H]
    \centering
    \includegraphics[width=\linewidth]{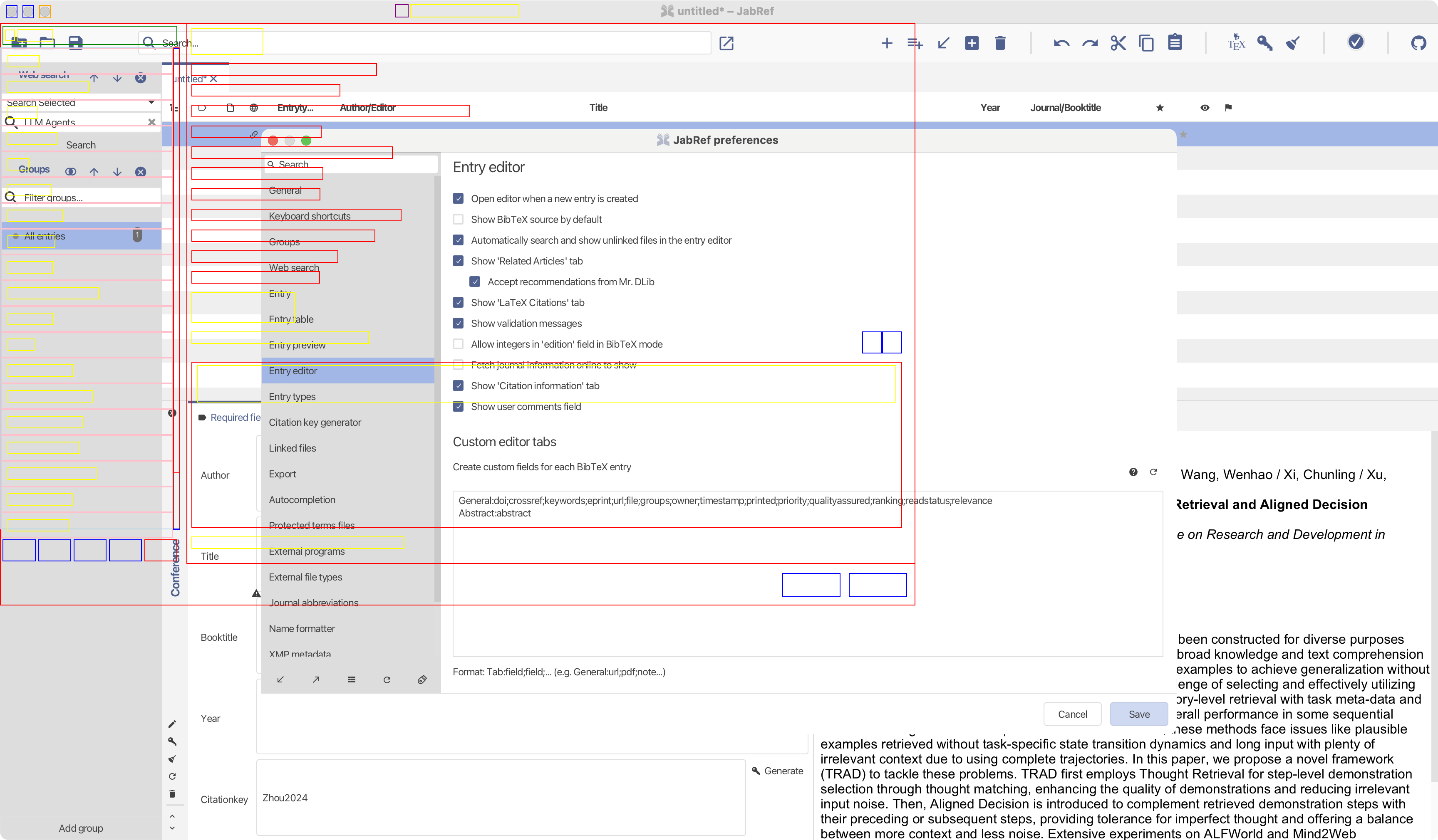}
    \Description{A screenshot of a macOS application where the bounding boxes provided by accessibility metadata are noticeably misaligned with the actual UI elements, affecting their usability.}
    \caption{Example of a macOS application with shifted UI element coordinates in the accessibility tree. The misalignment between reported and actual positions negatively impacts both assistive tools and agent-based interaction.}
    \label{fig:shifted-elements}
\end{figure}

\section{Accessibility issues}

macOS's built-in accessibility system, while powerful, often exhibits inconsistencies that can hinder both user experience and automated system performance. Two common issues we observed are invisible elements—elements listed in the accessibility metadata but not visually present on the screen—and shifted elements, where the reported coordinates do not align with their actual on-screen positions.
Figure~\ref{fig:invisible-elements} and Figure~\ref{fig:shifted-elements} illustrate these challenges with real examples from macOS applications.

